\documentclass[lettersize,journal]{IEEEtran} \usepackage{amsmath,amsfonts} 
\usepackage{algorithm} 
\usepackage{algorithmicx} 
\usepackage{array} \usepackage[caption=false,font=normalsize,labelfont=sf,textfont=sf]{subfig} 
\usepackage{textcomp} 
\usepackage{stfloats} 
\usepackage{url} 
\usepackage{verbatim} 
\usepackage{graphicx} 
\usepackage{cite} 
\usepackage{algpseudocode} 
\usepackage{multirow} 
\usepackage{amssymb} 
\usepackage{orcidlink}
\usepackage{svg}

\begin{document}

\title{Hybrid Task and Motion Planning with Reactive Collision Handling for Multi-Robot Disassembly of Complex Products: Application to EV Batteries}

\author{Abdelaziz Shaarawy, Cansu Erdogan, Rustam Stolkin, Alireza Rastegarpanah\,
\\

\thanks{A. Shaarawy, C. Erdogan and R. Stolkin are with Extreme Robotics  Laboratory, School of Metallurgy and Materials, University of Birmingham, Birmingham, B15 2TT U.K.}
\thanks{A. Rastegarpanah is with Extreme Robotics  Laboratory, School of Metallurgy and Materials, University of Birmingham, Birmingham, B15 2TT U.K., and also with Department of Applied Artificial Intelligence and Robotics, Aston University, Birmingham, B4 7ET, U.K.}

}

\markboth{Journal of \LaTeX\ Class Files,~Vol.~14, No.~8, August~2021}%
{Shell \MakeLowercase{\textit{et al.}}: A Sample Article Using IEEEtran.cls for IEEE Journals}

\maketitle
\begin{abstract}
This paper addresses the problem of multi-robot coordination for complex manipulation task sequences. We present a vision-driven task-and-motion planning (TAMP) framework for a real dual-agent platform that integrates task decomposition and allocation with a learning-based RRT planner. A GMM-informed motion planner is coupled with a hybrid safety layer that combines predictive collision checking in a MoveIt/FCL digital twin with reactive vision-based avoidance and replanning. This integration is challenging as the system jointly satisfies task precedence, geometric feasibility, dynamic obstacle avoidance, and dual-arm coordination constraints. The framework operates in closed loop by updating the remaining task sequence from repeated scene scans and completion-state tracking rather than executing a fixed open-loop plan. In EV battery disassembly experiments, compared with Default-RRTConnect under identical perception and task assignments, the proposed system reduces cumulative end-effector path length from 48.8 to 17.9~m ($-63.3\%$), improves makespan from 467.9 to 429.8~s ($-8.1\%$), and reduces swept volumes (R1: $0.583\rightarrow0.139\,\mathrm{m}^3$, R2: $0.696\rightarrow0.252\,\mathrm{m}^3$) and overlap ($0.064\rightarrow0.034\,\mathrm{m}^3$). These results show that combining predictive planning and reactive collision avoidance in a real dual-arm disassembly scenario improves motion compactness, safety, and scalability to broader multi-robot sequential manipulation tasks.
\end{abstract}


\section{Introduction}
Task planning enables autonomous robots to execute complex, multi-step operations under resource constraints by selecting and sequencing actions. When coupled with motion planning, it determines \emph{how} and \emph{when} to act amid environmental dynamics and hardware limits. In single-robot settings, learning- and optimisation-based approaches (e.g., ANFIS, GA, RL) improve allocation and ordering under uncertainty \cite{TEJER2024107300}. In multi-robot systems, continuous coordination, collision constraints, and shared resources make scheduling substantially more challenging. Electric-vehicle (EV) battery disassembly is an especially demanding, safety-critical domain where tasks are sequential and tightly coupled, fixtures and tolerances are strict, and the workspace is cluttered and dynamic \cite{WEGENER2014155,Wang2023ABatReSimAC}. In practice, the main difficulty is not merely combining task planning and motion planning, but making them operate reliably under perception noise, dual-arm coordination complexity, dynamic obstacles, and simultaneous task-level and geometric constraints.

This work presents a perception-driven TAMP architecture for coordinated multi-robot operation in dynamic environments (Fig.~\ref{fig:TAMP_overview}). Perception serves as the entry point to the pipeline and feeds four planning and execution layers: task decomposition, task allocation, offline predictive motion planning, and online reactive execution. Stereo RGB-D sensing with YOLOv8 provides object identities and poses; the task layer enforces precedence, accessibility, and tool-compatibility constraints. The motion layer combines a MoveIt~\cite{DBLP:journals/corr/ColemanSCC14}/FCL~\cite{6225337} digital twin for predictive collision checking with a reactive avoidance module during execution. In EV battery disassembly, this closed-loop architecture continuously updates both sequences and trajectories from scene feedback, enabling robust dual-robot cooperation rather than a fixed open-loop pipeline.

This paper demonstrates a closed-loop dual-arm system in which planning, perception, and execution remain tightly coupled on real hardware. Building on the GMM-informed RRT planner introduced in \cite{SHAARAWY2026103095}, this work embeds that planner within a vision-driven dual-arm TAMP framework \cite{robotics13050075} that continuously handles object-state updates, task allocation changes, inter-robot interference, and dynamic collision risks while preserving the regulatory disassembly sequence. Hence, the main contributions of this article are as follows:

\begin{itemize}
    \item Hybrid predictive/reactive collision handling: we combine predictive digital-twin collision checking with reactive avoidance and vision-driven replanning. This coupling supports dual manipulation in dynamic and scalable multi-robot settings.
    \item Closed-loop coupling between task and motion layers: the system re-scans the scene, tracks object-completion status, and updates the remaining lookup-table sequence during execution, rather than treating task planning and motion planning as fully decoupled stages.
    \item Real-world validation in EV battery disassembly: across the presented three case studies, the integrated framework yields more compact trajectories, lower swept-volume overlap, and improved execution efficiency under real-world uncertainty.
\end{itemize}

\IEEEpubidadjcol

\section{Related Work}
\label{RelatedWork}
Task-and-motion planning (TAMP) is central to multi–robot operation in shared workspaces: it must choose feasible action sequences, allocate limited resources, and coordinate agents while guaranteeing geometric and kinematic safety. The integration of symbolic task planning with continuous motion planning is widely recognised as a key enabler for autonomous execution in complex environments through logic–based representations and discrete decision mechanisms.
\subsection{Motion Planning}
Motion planning computes safe, efficient trajectories under kinematic limits and obstacles; in multi-robot settings, it must also coordinate agents for collision-free, synchronised execution. Sampling-based motion planning (SBMP) provides probabilistic completeness and efficient exploration in high-dimensional spaces \cite{508439,LaValle1998RapidlyexploringRT}. Integrating SBMP with TAMP links discrete task decomposition to continuous geometric trajectories and supports dynamic replanning when tasks or preconditions change \cite{inproceedings}. Empirical studies (e.g., bin-picking) report substantial variation across SBMP variants in time, success, and path quality, motivating task-aware selection and tuning \cite{9911132}. Guided SBMP adapts sampling to real-time perception for on-the-fly obstacle handling \cite{khanal2023guidedsamplingbasedmotionplanning}, while kinodynamic extensions such as iDb-RRT combine motion primitives with optimisation to satisfy dynamics and yield smoother, executable paths \cite{ortizharo2024idbrrtsamplingbasedkinodynamicmotion}.

Learning-based planners target adaptation under uncertainty: deep models show strong performance in dense obstacles with curated datasets \cite{9613554}, and surveys compare deterministic, heuristic, and learning-based methods across optimality, computation, and robustness \cite{machines11070722}. Recent DRL advances coordinate global target sequencing with local obstacle-aware navigation via hierarchy and curriculum/experience reuse, outperforming RRT*/PRM in success rate \cite{Fan_2023}. 
Reliable perception underpins both sampling- and learning-based planners. Probabilistic 3D maps such as OctoMap enable dynamic updates for moving obstacles \cite{hornung13auro}. Recent task-conditioned planners also motivate our use of a learned sampling prior: the recently introduced GMM-informed RRT planner in \cite{SHAARAWY2026103095} demonstrated planner-level benefits in constrained manipulation; the present paper studies the distinct problem of integrating that planner into a closed-loop dual-arm TAMP system.
\vspace{-1em}

\subsection{Task Planning}
Task planning determines action sequences and allocates resources (time, energy, computation) for complex, time-sensitive tasks. In single-robot settings, learning- and optimisation-based approaches (e.g., ANFIS, GA, RL) improve ordering and adaptation under uncertainty \cite{562456321,TEJER2024107300}. Hybrid models link symbolic reasoning with continuous control (e.g., affordance-guided wayfields) to connect abstract plans to real-time motion in cluttered, human-shared spaces \cite{8594492}. Data-driven methods, notably DRL, enable adaptation but incur heavy training cost and may struggle in highly dynamic scenes \cite{TEJER2024107300}. 

Structured knowledge offers robustness via ontology-based planning that fuses symbolic and geometric reasoning; these methods formalise manipulation tasks and respect spatial constraints but depend on predefined ontologies and can be brittle in unfamiliar environments \cite{8612805}. Moving beyond dyads, scaling to multi-robot teams introduces distribution and synchronisation issues. General-purpose planners cover logical constraints yet leave real-time responsiveness under uncertainty less explored \cite{10092974}. Heuristic and metaheuristic schedulers (e.g., simulated annealing, dynamic scheduling) trade solution quality for speed \cite{inbook,WANG2024104604}. Optimisation-based formulations (MILP/CP) improve allocation and conflict avoidance, while geometry-aware variants incorporate proximity and subtask dependencies; nevertheless, many approaches face limits in online adaptability and scalability in cluttered workcells \cite{FATEMIANARAKI2023102770,9636119,9013090,electronics12092131}.
\vspace{-1em}

\subsection{Integrated TAMP and Multi-Robot Coordination}
Integrated TAMP couples symbolic reasoning with geometry-aware motion to satisfy semantic preconditions, spatial constraints, and resource limits in shared workcells \cite{9681247,BERNARDO2023109345}. Recent efforts move toward distributed, feedback-coupled planning with feasibility-aware distribution, but often assume fixed uncertainty models or analyse failure recovery only partially \cite{10802695,10802307}. In multi-robot settings, the main challenge is preserving global task feasibility while accommodating robot-specific reachability, interference, and asynchronous execution. Our work targets this gap by coupling perception updates, task allocation, predictive motion planning, and reactive replanning within one execution loop on real hardware.
\vspace{-1em}

\subsection{Hybrid Predictive/Reactive Collision Avoidance}
Shared workspaces require both predictive and reactive safety mechanisms. Predictive collision checking through motion-planning frameworks such as MoveIt and proximity libraries such as FCL enables geometric validation before execution \cite{DBLP:journals/corr/ColemanSCC14,6225337}, while probabilistic 3D maps such as OctoMap support online updates for moving obstacles \cite{hornung13auro}. However, predictive checking alone is insufficient when obstacles or neighbouring robots move during execution. This motivates hybrid schemes in which a validated plan is continuously monitored and locally corrected or replanned when the scene changes. The present paper contributes such a hybrid predictive/reactive formulation in a real dual-arm disassembly cell, where perception updates, inter-robot distance monitoring, and replanning must operate jointly rather than as isolated modules.
\vspace{-1em}

\section{System Configuration}
\begin{figure}[!t]
\centering
\includegraphics[width=0.86\columnwidth]{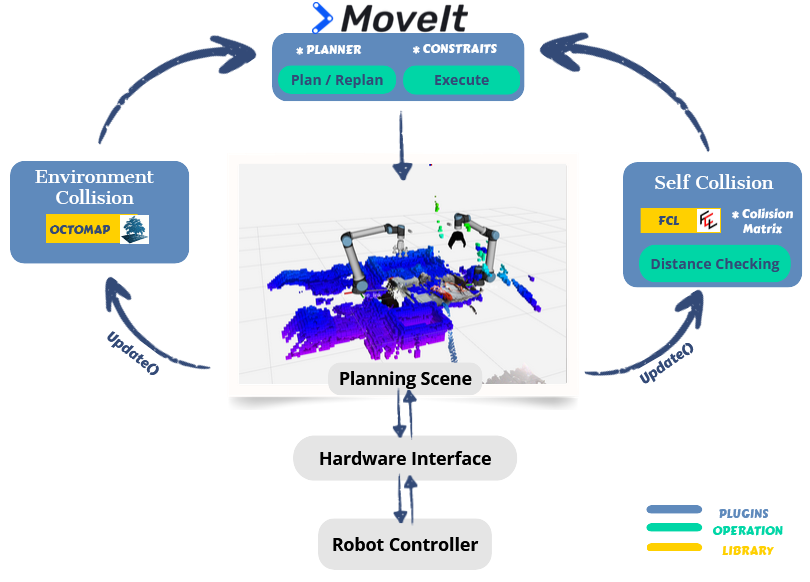}
\caption{System architecture for predictive planning and reactive execution in MoveIt.}
\label{system}
\vspace{-1.5em}
\end{figure}

We implement a ROS-based multi-robot stack for safe, coordinated operation in shared workspaces. A MoveIt-based control node orchestrates motion planning, manipulation, trajectory execution, and collision avoidance. The \emph{Planning Scene} aggregates robot kinematics, sensor input, and environment representations into a continuously updated state used by the planner (OMPL) and collision checker (FCL). Fig.~\ref{system} outlines the architecture and data flow.
 
\paragraph{MoveIt and the planning problem}
Planning and replanning are posed as a constrained optimisation over robot configurations $\mathbf{q}\in\mathcal{C}\subset\mathbb{R}^n$:
\[
\min_{\mathbf{q}\in\mathcal{C}} \; f(\mathbf{q}) \quad \text{s.t.} \quad d(\mathbf{q},t)\;\ge\; d_{\min},
\]
where $f(\mathbf{q})$ encodes path/time/energy costs, and $d(\mathbf{q},t)$ is the \emph{signed distance margin} to the nearest obstacle at time $t$ (positive: safe, zero: contact, negative: penetration). Safety is enforced \emph{along the trajectory}, i.e., $d(\mathbf{q}(s),t)\ge d_{\min}$ for all path parameters $s\in[0,1]$.

\paragraph{Environment collision}
An OctoMap-based 3D occupancy model provides geometry for environment queries. Let $d_{\mathrm{env}}(\mathbf{q},t)$ be the minimum signed distance from any robot link to mapped obstacles. The constraint
\[
d_{\mathrm{env}}(\mathbf{q},t) \;\ge\; d_{\min}
\]
ensures sufficient separation; whenever violated, the system triggers replanning or imposes local motion constraints to restore feasibility.

\paragraph{Self-collision}
Self-collision is monitored via FCL using a precomputed collision matrix and continuous distance checks between link pairs. With $d_{\mathrm{self}}(\mathbf{q})$ the minimum inter-link signed distance,
\[
d_{\mathrm{self}}(\mathbf{q}) \;\ge\; d_{\min}
\]
guarantees collision-free internal configurations during execution.

\paragraph{Planning Scene (constraint aggregation)}
The Planning Scene maintains both constraints concurrently; equivalently,
\[
d_{\mathrm{total}}(\mathbf{q},t)\;=\;\min\{\,d_{\mathrm{env}}(\mathbf{q},t),\; d_{\mathrm{self}}(\mathbf{q})\,\}\;\ge\; d_{\min}.
\]
This aggregation prevents false safety due to additive cancellation and supports consistent, real-time updates as perception refreshes the map and as robots move.

\paragraph{Robot Controller} 
The robot controller executes time-parameterised joint trajectories and streams setpoints to the drives with safety limits, interpolation, and controlled stopping on faults. Joint states and tracking errors are fed back to the planning stack, closing the loop between planning and physical actuation for synchronised dual-robot execution.

\vspace{-1em}

\section{Task-and-motion planning (TAMP)}
This section presents our task-and-motion planning (TAMP) framework for autonomous multi-robot disassembly of EV battery packs. Implemented in ROS/MoveIt, the system comprises (i) a logic-based task planner that schedules symbolic actions under temporal, spatial, and resource constraints and (ii) a learning-guided motion planner that produces adaptive, collision-free trajectories satisfying geometric and kinematic feasibility. Real-time perception, online replanning, and reactive control enable robust, coordinated, and safe behaviour in dynamic, unstructured environments.

\begin{figure*}[!t]
\centering
\includegraphics[width=0.94\textwidth]{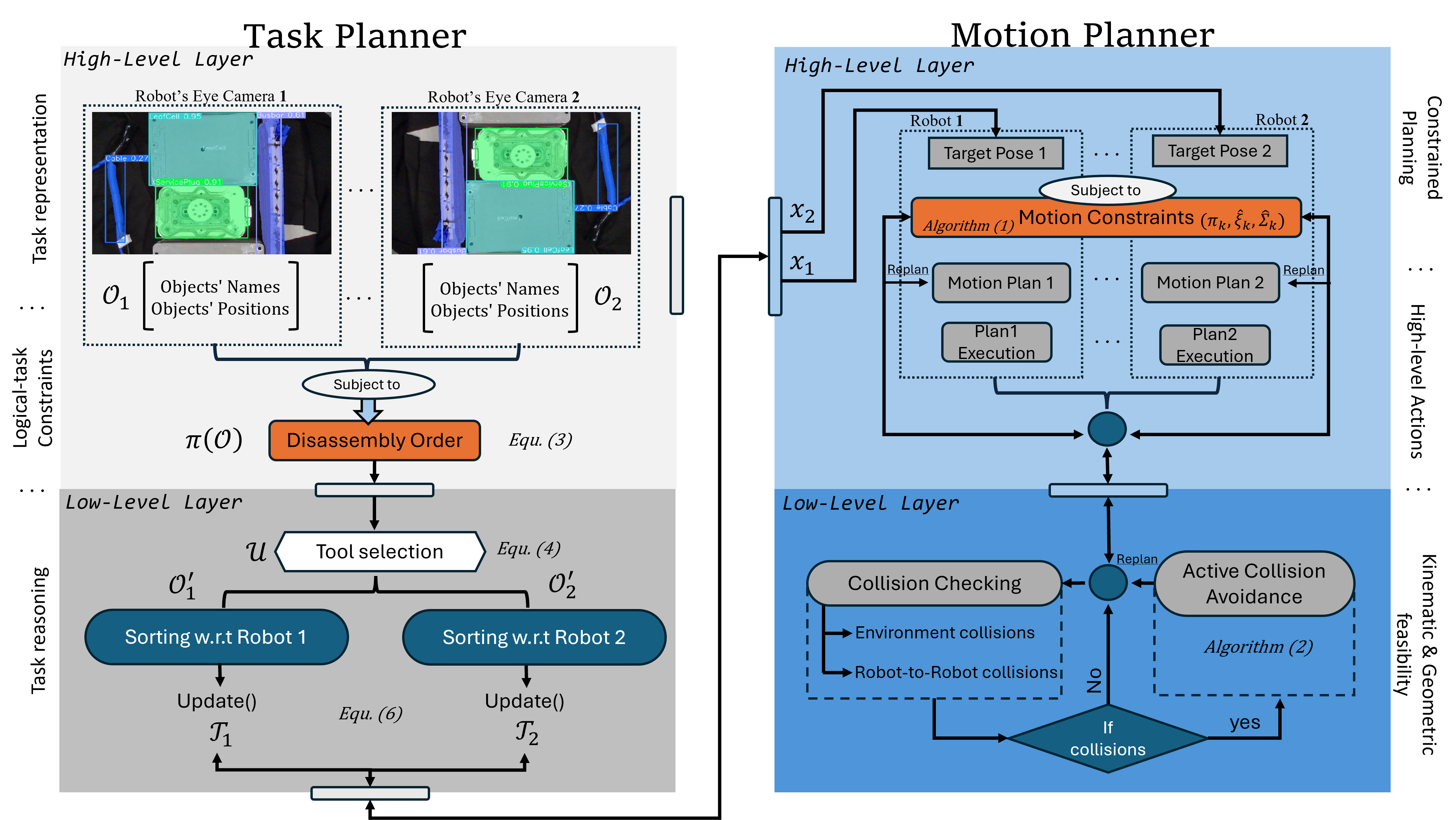}
\caption{Overview of the proposed TAMP framework in a dual-arm setup. Perception feeds task decomposition, task allocation, offline/predictive planning, and online/reactive execution in a closed loop.}
\label{fig:TAMP_overview}
\vspace{-1.5em}
\end{figure*}

\subsection{Task Planner}\label{sec:task_planner}


Fig.~\ref{fig:TAMP_overview} summarises the perception-driven task-planning stack that coordinates the disassembly with two manipulators. Perception is treated as an explicit front-end rather than as part of the task planner itself: Intel RealSense RGB-D sensing first provides component IDs and 3D pose estimates, which are then consumed by a \textit{task-decomposition} module to enforce precedence and by a \textit{task-allocation} module to assign robots and tools. Execution is monitored in closed loop; after each completed, aborted, or re-planned action, the robots reset to observation poses, the scene is re-scanned, and the remaining task sequence is updated while preserving safety constraints. Hence the task layer is coupled to motion and execution through repeated scene updates rather than operating as a one-shot symbolic plan.

\subsubsection{Perception and Task Representation}
The perception pipeline begins with a \textit{task representation} stage, where an Intel RealSense RGB-D camera is integrated with YOLOv8-Seg~\cite{YOLOV8} to detect and segment battery-pack components at the part level. Following the methodology in~\cite{robotics13050075}, but using real rather than simulated data, we report detection, segmentation, and 3D localisation metrics under real-world conditions. At the beginning of each execution cycle, and again after each completed or aborted manipulation, the active robot returns to its observation pose so that the current scene state can be verified before the next action is committed.  
A dataset of 226 real images of NISSAN e-NV200 battery components was collected, covering four object classes: cable, busbar, leaf cell, and service plug. Of these, 204 images were allocated for training and 22 for validation. Manual labelling and dataset augmentation expanded the dataset to 542 images (474 training, 45 validation, and 23 test). Model training was performed using YOLOv8 on Google Colab (Ubuntu 20.04, Intel Core i7-13700H CPU, NVIDIA GeForce RTX 3050 GPU). Input images were resized to $640 \times 640$ pixels, balancing accuracy and computational efficiency. Training was run for 50 epochs and completed in approximately 0.317 hours. The final trained model was exported and integrated into the perception stage.
The segmentation masks are fused with depth information to recover each component’s 3D centroid and, when required, its 6-DoF pose. These estimates are then provided to the grasping and motion planning modules for execution. After 6-DoF pose acquisition, the information is passed to the \textit{logical-task} stage, where objects are sorted into a predetermined disassembly sequence~\cite{10807203}. This ensures that task reasoning, safety checks, and robot/tool selection are performed in the correct order before execution proceeds.

\subsubsection{Multi-Robot Task Scheduling}
The \textit{Task Planner} module generates ordered task sequences for each robot, subject to logical and physical constraints. Let

\vspace{-1em}
\begin{equation}
\mathcal{O} = \{ o_1, o_2, \dots, o_N \}
\end{equation}
denote the set of detected disassembly components, where each component $o_i$ is represented as

\vspace{-1em}
\begin{equation}\label{eq:object_representation}
o_i = (label_i, n, \mathbf{x}_i, u_i, s_i),
\end{equation}
with class label $label_i$, number of detected instances $n$, $\mathbf{x}_i = { x_{i1}, \dots, x_{in} }$ the list of estimated object poses $x_{in} \in SE(3)$, required tool $u_i \in \mathcal{U}$, and completion state $s_i \in \{\texttt{pending},\texttt{in-progress},\texttt{completed}\}$.

\paragraph{Stage 1: Object Detection}  
At the initial state, and again after each execution update, each robot’s eye-in-hand RGB-D camera acquires a frame from a predefined observation pose and queries a YOLOv8-based detection server. The output is the current unordered set $\mathcal{O}$ of detected objects, their 3D poses, and their latest completion states.  

\paragraph{Stage 2: Priority-based Sorting}  \label{sec:lookup_table}
Each object $o_i$ is assigned a priority score $\pi(i)$ by consulting an expert-encoded lookup table that reflects safety, disassembly dependencies, accessibility, and tool constraints, following~\cite{10807203}. This lookup-table design is intentional: for EV battery disassembly, regulatory procedure and safety compliance are better expressed as hard expert rules than as soft costs inside a generic symbolic search. The table therefore encodes the abstract task structure of the process, while remaining extensible--new object classes or task types can be incorporated by adding new rows with precedence, tool, and safety annotations. Objects are then ordered according to the predefined disassembly sequence $\pi$,  
\vspace{-1em}
\begin{equation}
\label{eq:disassembly_order}
\mathcal{O}' = \pi(\mathcal{O})    
\end{equation}

ensuring logical ordering (e.g., screws $\prec$ battery module). Table~\ref{tab:prioritysorting} illustrates the mapping between priority scores, battery components, and required tools.

\begin{table}[ht]
\centering
\begin{tabular}{c c c} 
\hline
\textbf{Priority Score $(\pi)$} & \textbf{Component} & \textbf{Required Tool} \\ \hline
1 & Screws & 3F Gripper \\ 
2 & Battery Module & Vacuum Gripper \\ 
3 & Service Plug & 3F Gripper \\ 
4 & Busbar & 3F Gripper \\ 
5 & Cable & 3F Gripper \\ 
\hline
\end{tabular}
\caption{Expert-encoded lookup table for EV battery disassembly, mapping component classes to precedence priority and required tool. }
\label{tab:prioritysorting}
\vspace{-1.5em}
\end{table}

\paragraph{Stage 3: Tool-based Assignment}  
Let $\mathcal{R} = \{R_1, R_2\}$ denote the set of robots, each equipped with a distinct tool set $\mathcal{U}_j \subseteq \mathcal{U}$. The prioritized task list $\mathcal{O}'$ is partitioned into subsets
\begin{equation}
\mathcal{O}_j' = \{ o_i \in \mathcal{O}' \mid u_i \in \mathcal{U}_j \},
\end{equation}
such that each task $o_i$ is assigned to the robot $R_j$ capable of executing it.  

\paragraph{Stage 4: Cost-based Sorting}  
After logical ordering and tool assignment, objects in each subset are further sorted according to an execution cost function $\mathcal{J}_j$. Generally, $\mathcal{J}_j$ can encode various optimisation criteria (e.g., manipulability, collision risk, energy). In this work, we define it as a distance-based cost. For each robot $R_j$, let $\mathbf{p}_j \in SE(3)$ denote its current end-effector pose. The execution cost is defined as:

\vspace{-1em}
\begin{equation}\label{eq:distance_cost}
    \begin{aligned}
    \mathcal{J}_j &= min \sum_{i=1}^N d(\mathbf{x}_i, \mathbf{p}_j), \\
    d(\mathbf{x}_i, \mathbf{p}_j) &= | \mathbf{x}_i - \mathbf{p}_j |_2 
    \end{aligned}
\end{equation}

The final ordered task sequence for $R_j$ is obtained by solving

\vspace{-1em}
\begin{equation}
\mathcal{T}_j = \arg\min_{\mathbf{x}} \sum_{i=1}^{|\mathcal{O}_j'|} d(\mathbf{x}_i, \mathbf{p}_j),
\end{equation}

where $|\mathcal{O}'_j|$ is number of components detected ($labels$) in the prioritized task list $\mathcal{O}'_j$ for $R_j$. Finally, the Task Planner outputs  
\[
\mathcal{T} = \{ \mathcal{T}_1, \mathcal{T}_2 \},
\]
where each $\mathcal{T}_j$ is an optimized task sequence for robot $R_j$. These sequences respect (i) disassembly precedence, (ii) tool compatibility, and (iii) distance-based efficiency, and are passed to the \textit{Motion Planner} for trajectory generation.
\paragraph{Closed-Loop Task-State Update}
After each pick, place, or interruption event, the robots reset to their observation poses and re-scan the workspace to verify the current object states. If an item has been successfully removed from its source and confirmed at its destination, its completion state is set to \texttt{completed}; otherwise it remains \texttt{pending} or is returned to the queue for replanning. The lookup-table ordering is then updated by removing completed objects and re-sorting the remaining items according to the refreshed scene, accessibility, and tool constraints. Consequently, task decomposition and task allocation are updated online as the motion layer changes the scene, showing that the two layers are coupled in closed loop rather than fully decoupled.
\vspace{-1.5em}

\subsection{Predictive Motion Planning and Reactive Execution}
In this section, we describe the motion component of the proposed TAMP framework for multi-robot setups. For each robot $R_j$, the planner receives the optimized task sequence $\mathcal{T}_j$ from the task layer together with the latest scene state. The corresponding object poses $\mathbf{x}_i$ (from Equation~\ref{eq:object_representation}) are extracted and transformed from the camera coordinate frame into the robot’s reference frame. Based on these task targets, the motion layer computes feasible trajectories by addressing two key requirements: (i) offline/predictive path generation, achieved via a Learning-from-Demonstration (LfD) approach that biases the sampling process toward expert-like trajectories, and (ii) online/reactive collision handling, ensuring safe execution in multi-robot and cluttered environments.

\subsubsection{LfD-Guided Motion Planning}\label{sec:lfd-guided_planner}
\hfill

As demonstrated in \cite{SHAARAWY2026103095}, integrating Learning-from-Demonstration (LfD) techniques with sampling-based motion planning algorithms leads to highly efficient motion planners. That prior work reported the planner-level comparisons and algorithmic evaluation of the GMM-informed RRT module itself; the present paper therefore builds on that foundation and focuses on integrating the validated planner into a vision-driven dual-arm TAMP system with closed-loop task updates and reactive collision handling. The motion-planning pipeline begins with the collection of demonstrations. Following the approach in~\cite{TeleOp_paper}, demonstrations were recorded for a set of disassembly tasks, extended in this work to a dual-robot setup. In these sessions, one robot executed random motions within the shared workspace, while the other was teleoperated by a human expert. The demonstrations encode trajectories from start to goal poses while avoiding collisions with the other robot. These data capture expert approaches for safe and efficient manipulation in multi-robot environments and are subsequently used to train a probabilistic model that generalises the underlying motion patterns of both robots.

\paragraph{TP-GMM Learning and Reproduction}
Gaussian Mixture Models (GMMs) are employed to capture the variability of demonstrated trajectories in Cartesian space, representing motion data as a weighted sum of $K$ Gaussian components. Model parameters are estimated using the Expectation-Maximisation (EM) algorithm, enabling probabilistic generalisation of expert demonstrations. However, a standard GMM is restricted to a fixed reference frame, which limits its adaptability to changing task conditions. 
Task-Parameterised GMMs (TP-GMMs) overcome this limitation by encoding demonstrations with respect to multiple task-dependent frames. Following the approach in \cite{SHAARAWY2026103095}, these frames are denoted as the start pose $\hat{f}^{(s)}$, goal pose $\hat{f}^{(g)}$, and obstacle poses $\hat{f}^{(o)}$, corresponding to $\mathbf{p}_{k}$ and $\mathbf{x}_{ik}$ (Equation \ref{eq:distance_cost}), all expressed relative to the robot’s $R_j$ base frame. 
Since this work extends \cite{SHAARAWY2026103095} to a dual-robot setup, for each defined task (e.g., pick or place) two TP-GMMs are learned, one per robot. Now each task demonstration is expressed in the corresponding frames, yielding three GMMs that are subsequently combined. During reproduction, new task parameters are provided, and the model adapts its Gaussian components as follows:
\begin{equation} \label{eq:reproduced_gmm}
    \begin{aligned}
    \boldsymbol{\hat{\xi}}^{(s)}_{t,k} &= \hat{A}^{(s)}_{t} \, \boldsymbol{\mu}^{(s)}_k + \hat{b}^{(s)}_{t}, \quad 
    \boldsymbol{\hat{\Sigma}}^{(s)}_{t,k} = \hat{A}^{(s)}_{t} \, \boldsymbol{\Sigma}^{(s)}_k \, (\hat{A}^{(s)}_{t})^T \\
    \boldsymbol{\hat{\xi}}^{(g)}_{t,k} &= \hat{A}^{(g)}_{t} \, \boldsymbol{\mu}^{(g)}_k + \hat{b}^{(g)}_{t}, \quad 
    \boldsymbol{\hat{\Sigma}}^{(g)}_{t,k} = \hat{A}^{(g)}_{t} \, \boldsymbol{\Sigma}^{(g)}_k \, (\hat{A}^{(g)}_{t})^T \\
    \boldsymbol{\hat{\xi}}^{(o)}_{t,k} &= \hat{A}^{(o)}_{t} \, \boldsymbol{\mu}^{(o)}_k + \hat{b}^{(o)}_{t}, \quad 
    \boldsymbol{\hat{\Sigma}}^{(o)}_{t,k} = \hat{A}^{(o)}_{t} \, \boldsymbol{\Sigma}^{(o)}_k \, (\hat{A}^{(o)}_{t})^T
    \end{aligned}
\end{equation}

The number of Gaussian components is set to $K=5$, as determined by the Bayesian Information Criterion (BIC) in \cite{SHAARAWY2026103095}. This results in an adapted GMM with updated parameters 
$\{ \pi_k, \boldsymbol{\hat{\xi}}_{t,k}, \boldsymbol{\hat{\Sigma}}_{t,k} \}_{k=1}^{K}$, 
which generates trajectories consistent with the demonstrated behaviours while conditioned on the new task configuration.

\paragraph{GMM-Informed Planner}
Our planning framework integrates task space and configuration space. Task space, a subset of $SE(3)$, represents object poses, grasping poses, and manipulation goals, while configuration space $\mathcal{C} \subseteq \mathbb{R}^n$ handles the robot’s joint states. The TP-GMM is trained and reproduced in task space, encoding expert demonstrations relative to task parameters' frames ($\hat{f}^{(s)}, \hat{f}^{(g)}, \hat{f}^{(o)}$, and then produce an adapted distribution $\{\pi_k, \boldsymbol{\hat{\xi}}_{t,k}, \boldsymbol{\hat{\Sigma}}_{t,k}\}_{k=1}^{K}$. This defines a high-likelihood region $\mathcal{Q}$ in Cartesian space, which is enforced as a constraint during planning.
Path generation is performed using the RRTConnect (also referred to as Bi-RRT) algorithm from OMPL \cite{sucan2012the-open-motion-planning-library}, where $\mathbf{q}_s$ (the robot’s current configuration) initializes the start tree $Tree_s$, and $\mathbf{q}_g$ is computed via inverse kinematics of $\hat{f}^{(g)}$. At each iteration, a candidate configuration $\mathbf{q}_{sample} \in \mathcal{C}$ is drawn. Through rejection sampling, only configurations whose forward kinematics satisfy $FK(\mathbf{q}_{sample}) \in \mathcal{Q}$ are retained. This biases exploration toward task-relevant regions, guiding the planner toward expert-like motions.  
Algorithm \ref{alg:rrt_connect} outlines the GMM-constrained RRTConnect introduced in \cite{SHAARAWY2026103095}. The validated samples are used to extend the bidirectional trees, and once connected, a feasible path $\mathcal{X} \in \mathcal{Q}$ is returned. Trajectory $\mathcal{X}$ is now a valid, collision-free solution that accomplishes the primary pick-object task while satisfying configuration-space (kinematic) constraints, as well as remaining consistent with expert demonstrations, thus satisfying task-space constraints.


\begin{algorithm}[t]
\small
\caption{GMM-Informed Bi-RRT}
\label{alg:rrt_connect}
\begin{algorithmic}[1]

\Require Start, $\hat{f}^{(s)}$, Goal $\hat{f}^{(g)}$, and Obstacle $\hat{f}^{(o)}$ task pose frames
\Ensure A feasible Cartesian path $\mathcal{X}$ from $\hat{f}^{(s)}$ to $\hat{f}^{(g)}$ : $\mathcal{X} \in \mathcal{Q}$

\State $\mathcal{Q} \gets$ reproduceGMM($\hat{f}^{(s)}$, $\hat{f}^{(g)}$, $\hat{f}^{(o)}$) \Comment{(eq. \ref{eq:reproduced_gmm}})
 
\State $q_s \gets q_{current}$ (robot's current configuration)
\State $q_g \gets IK(\hat{f}^{(g)})$
\State $Tree_s.$InitTree($q_{\text{s}}$)
\State $Tree_g.$InitTree($q_{\text{g}}$)

\While{not Terminated()}
    \State $q_{\text{sample}} \gets$ SampleConfiguration($\mathcal{Q}$)
        \State $q_{\text{new}} \gets$ \Call{ExtendTree}{$Tree_s$, $q_{\text{sample}}$}
        \If{$q_{\text{new}} \neq$ None}
            \If{\Call{ConnectTree}{$Tree_g, q_{\text{new}}$}}
                \State \Return \Call{ExtractPath}{$Tree_s$, $Tree_g$}
            \EndIf
        \EndIf
    \State \Call{Swap}{$Tree_s$, $Tree_g$}
\EndWhile
\State \Return \textsc{Failure}
\end{algorithmic}
\end{algorithm}

\subsubsection{Reactive Collision Monitoring and Avoidance}
\hfill

In addition to path optimisation through LfD, the second critical component of our motion planner is \textit{collision perception and avoidance}, which ensures safe execution of trajectories in a shared dual-arm workspace. We investigate two UR10e manipulators operating in proximity, where inter-robot collisions and interactions with dynamic obstacles are key challenges. To address this, the proposed Active Collision Avoidance module combines real-time perception with avoidance strategies, structured into two complementary parts: \textit{Perception} and \textit{Avoidance}.

\paragraph{\textit{Collision Perception}}
Collision detection is handled through the Flexible Collision Library (FCL) within a MoveIt Planning Scene. All collision objects, including the two robots (from their URDFs) and the environment, are monitored in real time.

\begin{figure}[ht]
\centering
\includegraphics[width=\columnwidth]{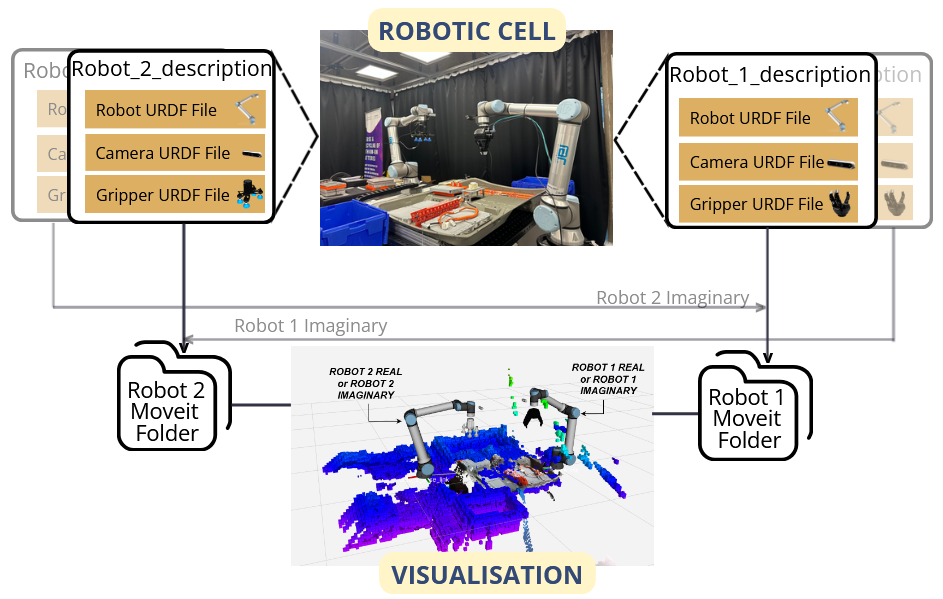}
\caption{Integration of real and virtual (imaginary) robots in a shared simulation, where each URDF embeds the counterpart’s model to enable inter-robot collision checking through FCL.} 
\label{URDF}
\vspace{-1em}
\end{figure}

\textsc{Robot-to-Robot Collisions:}
MoveIt’s Allowed Collision Matrix (ACM) is used to encode feasible body interactions. Since dual-robot systems require bidirectional awareness, we employ a digital twin strategy: each URDF embeds a replica of the other robot (Fig.~\ref{URDF}), enabling independent but coordinated collision checking. The ACM is further refined by excluding irrelevant body pairs to reduce computational load. 
\vspace{-1.5em}

\textsc{Environment Collisions:}
An Intel RealSense depth camera provides dynamic 3D workspace perception via MoveIt’s Occupancy Map Updater, which converts depth data into a voxel-based OctoMap with 0.03~m resolution. This OctoMap is broadcast to both robots’ planning scenes, allowing each to reason about environmental constraints in real time. Due to the computational demands of maintaining two planning scenes, the maximum achievable update rate is $\sim$10~Hz.

\begin{figure}[!t]
\centering
\includegraphics[width=0.88\columnwidth]{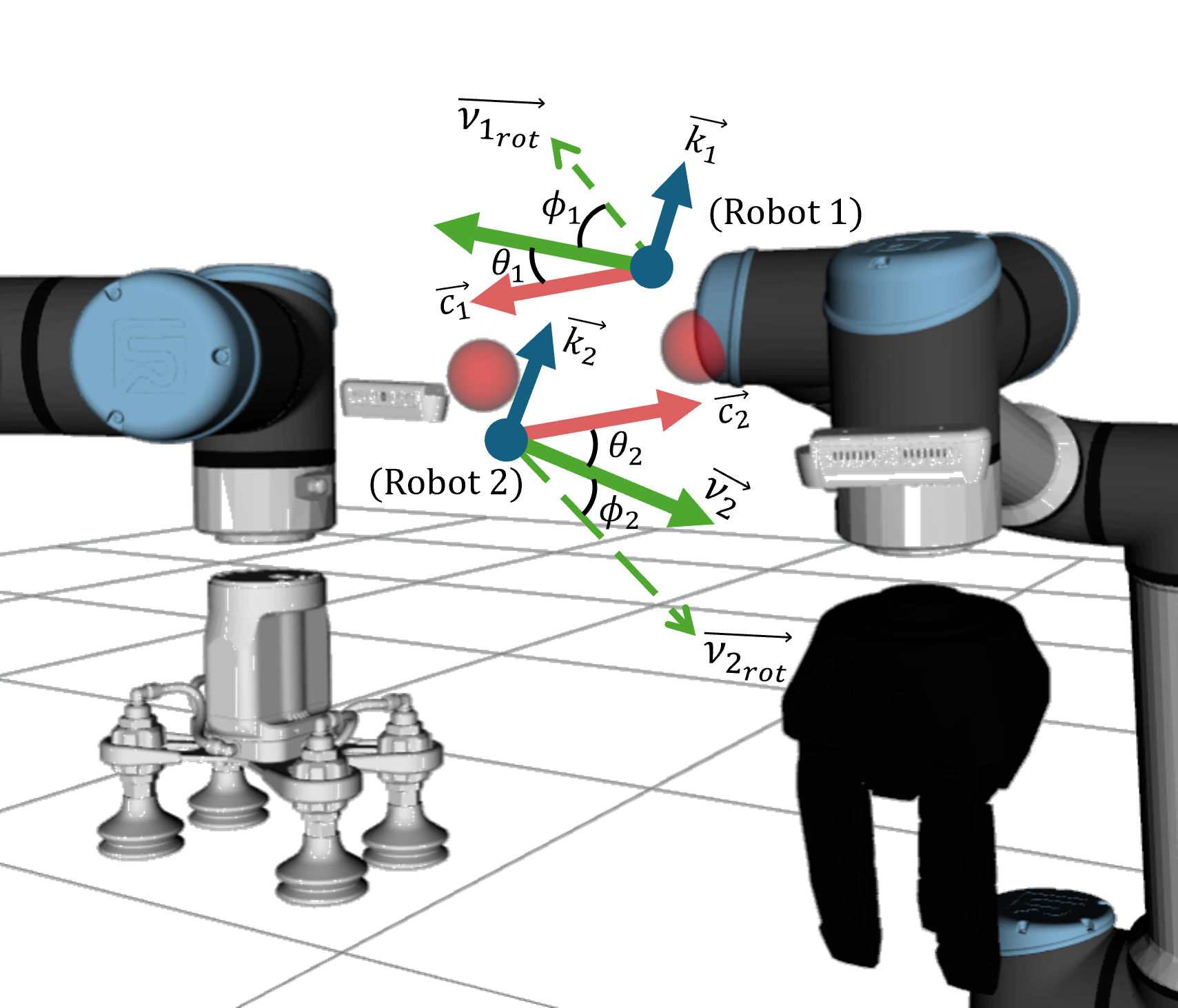}
\caption{Reactive \texttt{avoid\_collision()} routine triggered when two robots approach the early-warning threshold $\varepsilon$. The red spheres denote the contact pair reported by FCL, after which the controller locally steers away and requests replanning.} 
\label{fig:active_collision_avoidance}
\vspace{-1.5em}
\end{figure}

A safety threshold $\varepsilon = 0.15$~m is defined to detect near-collisions, with violations triggering reactive adjustments to increase separation. This corresponds to five OctoMap voxels and was chosen as a conservative early-warning margin that provides reaction distance at the available update rate while avoiding excessive false triggers. Limiting collision checks to distances $\leq \varepsilon$ reduces planning complexity. When a proximity violation is detected, FCL returns 3D contact points between objects, which are visualised in RViz (\texttt{visualization\_msgs/MarkerArray}) as shown in Fig. \ref{fig:active_collision_avoidance}.

\begin{algorithm}[ht]
\small
\caption{Reactive Collision Monitoring and Replanning}\label{alg:active_collision}
\begin{algorithmic}[1]
    \State Initialise planning-scene monitor and collision/distance services
\While{system active}
    \State Update planning scene from sensor and robot-state feedback
    \State Query collision state and minimum distance
    \If{collision detected}
        \State Stop current trajectory execution
    \ElsIf{minimum distance $\leq \varepsilon$}
        \State Invoke \texttt{avoid\_collision()} and request \texttt{replan()}
    \EndIf
    \State Publish collision and near-collision markers
\EndWhile
\end{algorithmic}
\end{algorithm}

\paragraph{\textit{Collision Avoidance}}  
Collision avoidance is implemented through a dedicated \texttt{avoid\_collision()} routine, enabling both robots to adapt dynamically to near-collisions in real time. The approach employs a velocity vector steering strategy in $SO(3)$, where the robot’s tool velocity vector $\vec{\mathbf{v}}$ is reoriented when a collision vector $\vec{\mathbf{c}} = \sum_{n=1}^{N}\vec{c_n}$ (aggregated from detected near-collisions) indicates a potential obstacle (Fig.~\ref{fig:active_collision_avoidance}). The steering axis is computed as $\vec{\mathbf{k}} = \vec{\mathbf{c}} \times \vec{\mathbf{v}}$, and Rodrigues’ rotation formula is applied to yield a rotated velocity vector $\vec{\mathbf{v}}_{\text{rot}}$.  
\begin{equation}\label{eq:v_rot}
    \vec{\mathbf{v}}_{rot} = \vec{\mathbf{v}} + sin(\phi) \cdot (\vec{k} \times \vec{\mathbf{v}}) + (1-cos(\phi)) \cdot \vec{k} \times (\vec{k} \times \vec{\mathbf{v}})
\end{equation}
A steering angle $\phi$ regulates the correction magnitude. Rather than fixing $\phi$, we define it to be relative to the angle $\theta$ between $\vec{\mathbf{c}}$ and $\vec{\mathbf{v}}$:  

\vspace{-2em}
\begin{equation}\label{eq:steering_angle}
\theta = \cos^{-1}\left(\frac{\vec{\mathbf{c}} \cdot \vec{\mathbf{v}}}{\|\vec{\mathbf{c}}\| \|\vec{\mathbf{v}}\|}\right), 
\quad \phi = \frac{\pi}{2} - \theta,
\end{equation}
so that imminent collisions (small $\theta$) yield larger steering corrections, while tangential encounters (large $\theta$) produce minimal deviations.  

This adaptive behaviour avoids unnecessary trajectory deviations, helping to maintain path optimality while ensuring safety through timely avoidance manoeuvres. Algorithm \ref{alg:active_collision} presents the overall Active Collision Avoidance component in the motion planner, with \texttt{avoid\_collision} routine being invoked when \texttt{near-collision} state is flagged to interrupt the robot controller's execution of the current planned trajectory. The routine keeps in operation until the \texttt{near-collision} state is changed. Each robot now requests a \texttt{replan} (Algorithm \ref{alg:rrt_connect}) a collision-free trajectory to the first given target pose from the current pose/configuration. The Active Collision Avoidance algorithm continues to run iteratively until convergence to the target pose is achieved.
\vspace{-1.5em}

\subsection {Framework Overview and Information Flow (Dual-Arm Setup)}

Fig. \ref{fig:TAMP_overview} illustrates the proposed TAMP framework implemented on a dual-arm robotic setup, asynchronously controlling two robots to complete a full disassembly task. The framework is organised as a perception front-end plus four planning/execution layers, ensuring both long-horizon task feasibility and safe, coordinated dual-arm manipulation. The information flow proceeds as follows:

\paragraph{Perception Front-End}
Two eye-in-hand cameras mounted on the robots acquire RGB-D images of the shared workspace. Each camera detects relevant objects, yielding their names, six-dimensional poses, and current completion states, which are passed into the planning stack.

\paragraph{Task Decomposition}
The current scene state is mapped to an abstract disassembly state that respects precedence constraints and the expert lookup table.

\paragraph{Task Allocation}
Feasible operations are assigned to each robot according to tool compatibility, reachability, and execution cost.

\paragraph{Offline / Predictive Motion Planning}
Task allocations are translated into motion-planning goals for each robot, and the GMM-informed RRT planner generates predictive motion plans validated against the digital twin before execution.

\paragraph{Online / Reactive Execution}
The execution layer ensures safe motion through collision monitoring, active avoidance, and replanning. When a near-collision is detected, the \texttt{avoid\_collision()} routine is invoked (Equations~\ref{eq:v_rot}--\ref{eq:steering_angle}), followed by replanning from the current robot configuration to the target pose. After each action, the robots re-scan the scene before the next task is confirmed.

Therefore, the complete TAMP pipeline operates as a hierarchical loop:
\begin{enumerate}
    \item Perception update $\rightarrow$ task decomposition computes the current feasible disassembly order.
    \item Task allocation $\rightarrow$ predictive motion planning generates task-constrained trajectories.
    \item Reactive execution $\rightarrow$ if a proximity violation occurs, invoke active avoidance and replan.
    \item Re-scan the scene, update completion states, and repeat until all disassembly goals are achieved.
\end{enumerate}
\vspace{-1.5em}
\section{Experimental Study} \label{sec:exp_study}
Three case studies were conducted to evaluate the proposed multi-robot Task and Motion Planning (TAMP) framework. These are referred to as \textit{Case Study I}, \textit{Case Study II}, and \textit{Case Study III}. \textit{Case Study I} investigates robot-to-robot interactions, focusing on potential collisions and the robots’ ability to replan trajectories while executing tasks in a shared workspace. \textit{Case Study II} evaluates robot performance in the presence of both static and dynamic obstacles. Finally, \textit{Case Study III} integrates both aspects into a real-world application: EV battery disassembly, analysing coordination, and executing paths. Collectively, these case studies validate the robustness of the proposed approach under diverse operational conditions. 

\textbf{System Calibration:} All experiments were carried out using two UR10e cobots calibrated to a fixed world frame. Each robot was equipped with an eye-in-hand Intel RealSense camera, and an additional eye-to-hand camera was fixed in the workspace. The robots used distinct end-effector tools (Robot~1: 3-fingered gripper; Robot~2: vacuum gripper) as shown in Fig.~\ref{case_study_3}. The framework relies on a digital twin with continuous sim-to-real and real-to-sim synchronisation so that planned trajectories can be executed on hardware while sensor feedback updates the scene model.
\vspace{-1em}
\subsection{Case Study I}
\begin{figure}[!t]
  \centering
    \subfloat[\centering ]{\label{fig:case_study_1a}\includegraphics[width=0.92\columnwidth]{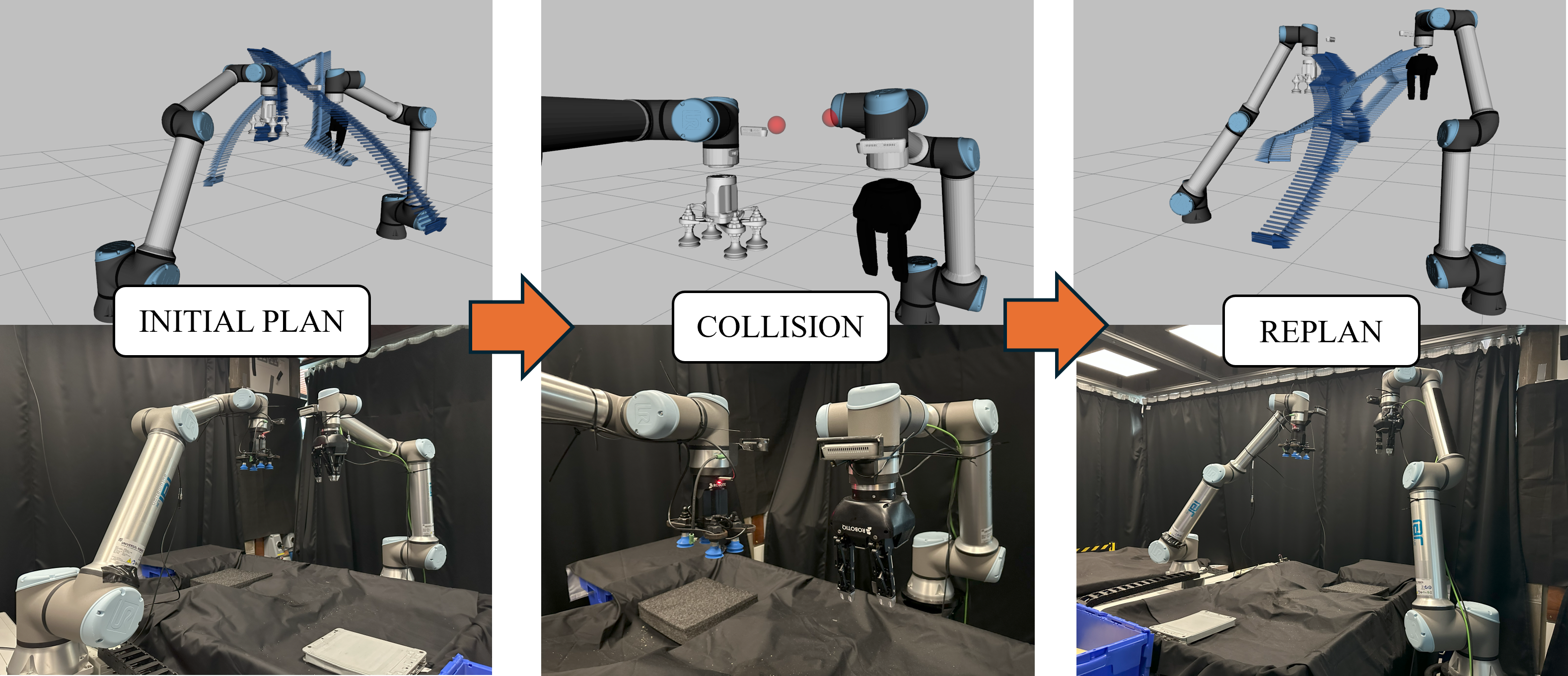}}
    \vspace{-1em}
    \subfloat[\centering]{\label{fig:case_study_1b}\includegraphics[width=0.92\columnwidth]{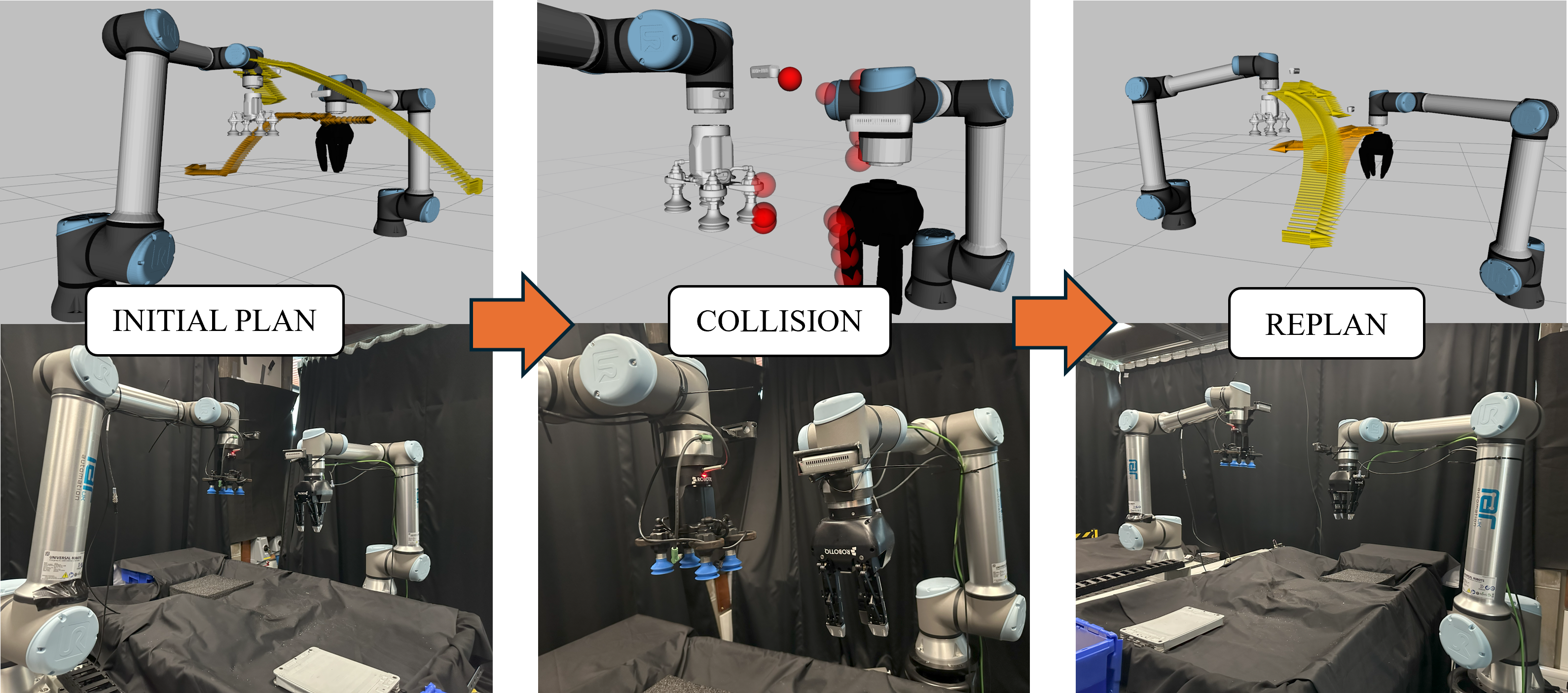}}\\
  \caption{Case study I: Robot-to-robot interaction. Comparison between (a) Default RRTConnect and (b) GMM-Informed RRTConnect in terms of planned-path quality. The figure displays the sequence in which an initial plan encounters a near-collision, triggering replanning that enables both robots to successfully reach their goal poses.}
  \label{fig:case_study_1}
  \vspace{-1.5em}
\end{figure}

\textit{Case Study I} evaluates the reliability of the motion layer within the TAMP framework in a dual-arm setup. Each robot is assigned a set of target poses corresponding to task allocations defined by the task layer. The focus is on comparing the LfD-guided GMM-informed RRTConnect (Section~\ref{sec:lfd-guided_planner}) against the baseline Uniform sampler RRTConnect. The objective is to assess the efficiency of both planners in generating feasible and optimal trajectories while avoiding robot-to-robot collisions in a shared workspace. 
Fig.~\ref{fig:case_study_1} illustrates the evaluation scenario, where both planners handle situations with potential inter-robot collisions. The study highlights each planner’s capacity to adapt to unpredictable workspace interactions and maintain safe navigation. 
\vspace{-1em}

\subsection{Case Study II}
\textit{Case Study II} evaluates the robots’ collision avoidance capabilities in environments with dynamic obstacles. The experimental setup comprises two UR10e robots performing collaborative tasks. To emulate dynamic disturbances, an object was attached to Robot~1’s end-effector and moved randomly within the workspace, while Robot~2 executed a transport task: picking up an object, carrying it, and placing it in a designated target area.  
\begin{figure}[!t]
  \centering
    \subfloat[\centering ]{\label{case_study_2_1}\includegraphics[width=0.92\columnwidth]{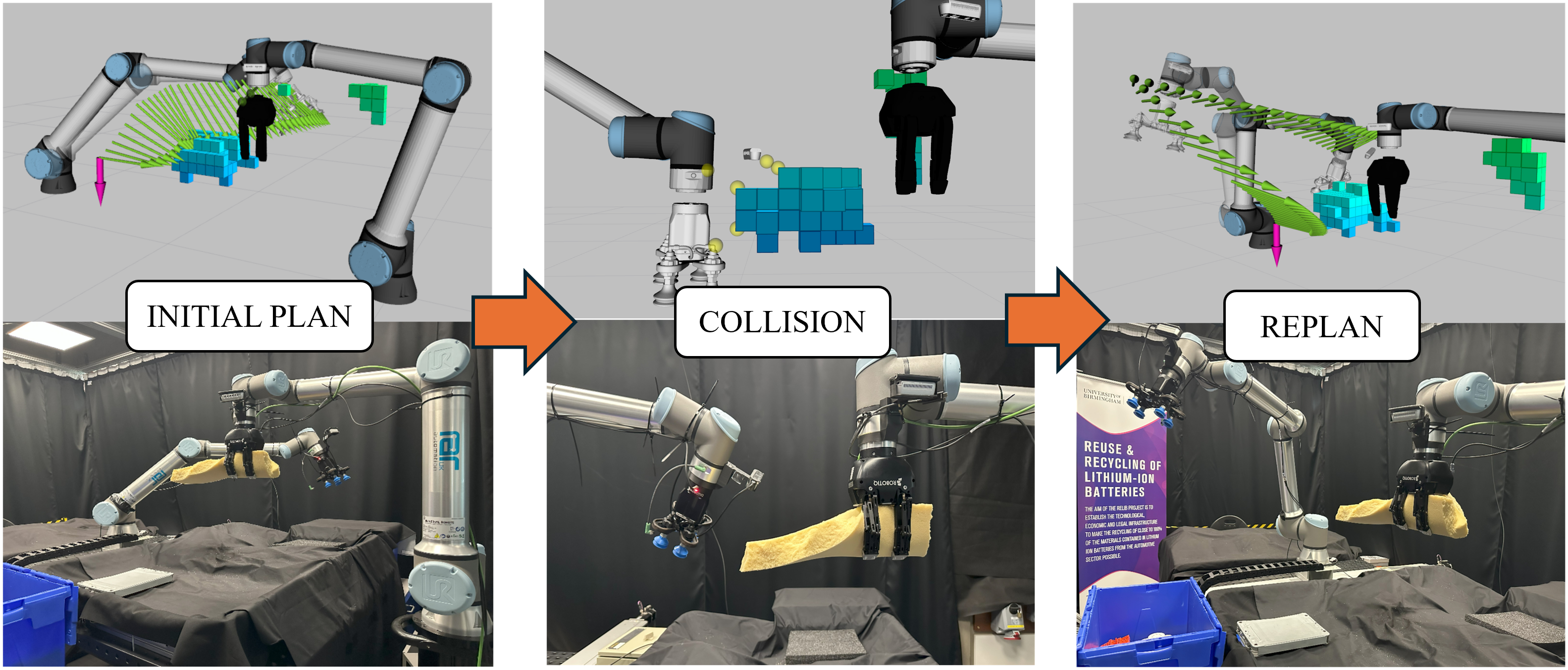}}
    \vspace{-1em}
    \subfloat[\centering]{\label{case_study_2_2}\includegraphics[width=0.92\columnwidth]{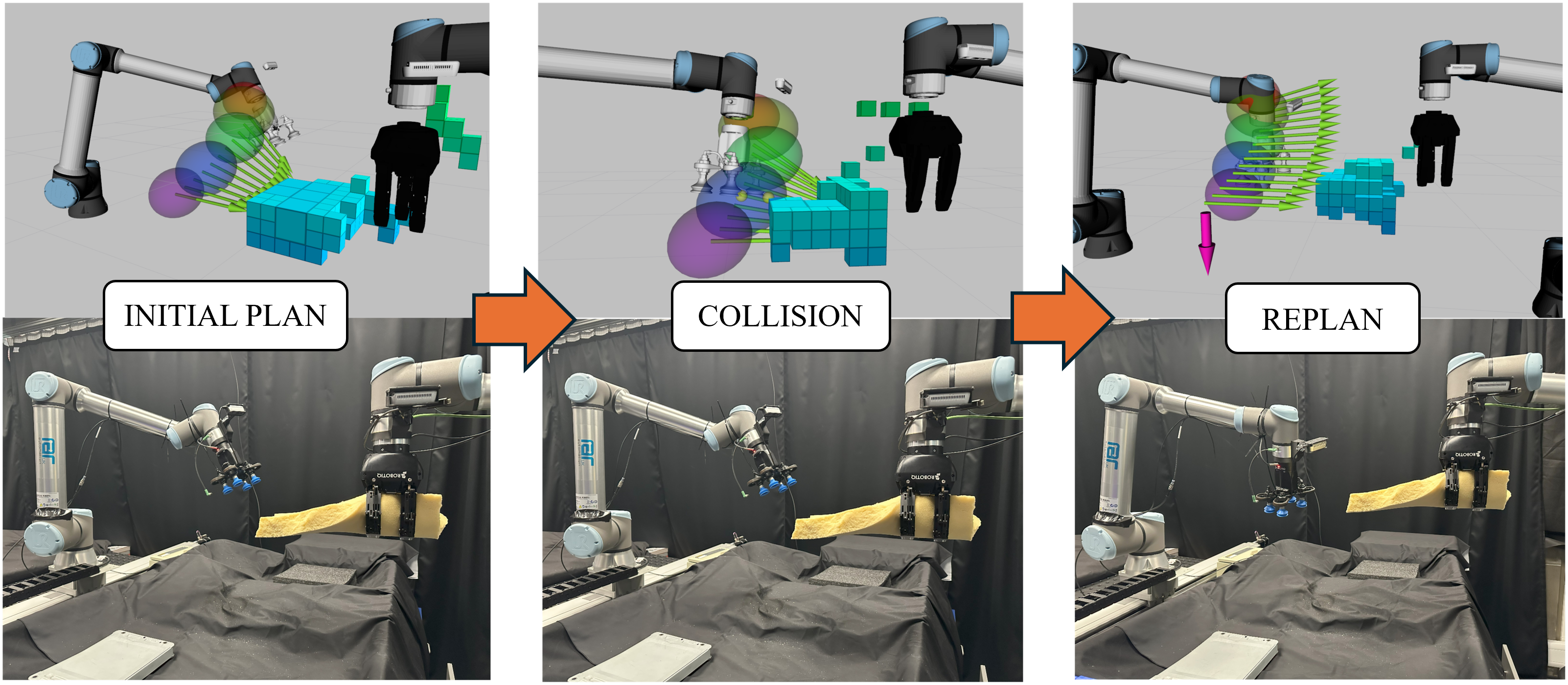}}\\
  \caption{Case Study II: Robot-to-dynamic environment interaction. Comparison between (a) Default RRTConnect and (b) GMM-Informed RRTConnect in terms of path quality. The figure illustrates the planning sequence where an initial plan encounters a near-collision with a dynamic object (in this case, attached to the other robot), triggering replanning that enables the robot arm to successfully reach its goal pose.}
  \label{fig:case_study_2}
  \vspace{-1.5em}
\end{figure}

Fig.~\ref{fig:case_study_2} shows the test scenario comparing Default RRTConnect and GMM-Informed RRTConnect. The sequence proceeds as follows: (Step 1) Robot~2 follows its initial plan toward the goal, (Step 2) a near-collision occurs due to Robot~1’s random motion, and (Step 3) Robot~2 invokes dynamic replanning to generate an alternative, collision-free path. This process ensures safe task completion while adapting to the dynamic environment. The study demonstrates how the proposed planner effectively handles robot-to-dynamic-environment interactions by eliminating collision risks while preserving task execution. 
\vspace{-1em}

\subsection{Case Study III}
\label{case_study_3_title}
\begin{figure}[!t]
\centering
\includegraphics[width=0.94\columnwidth]{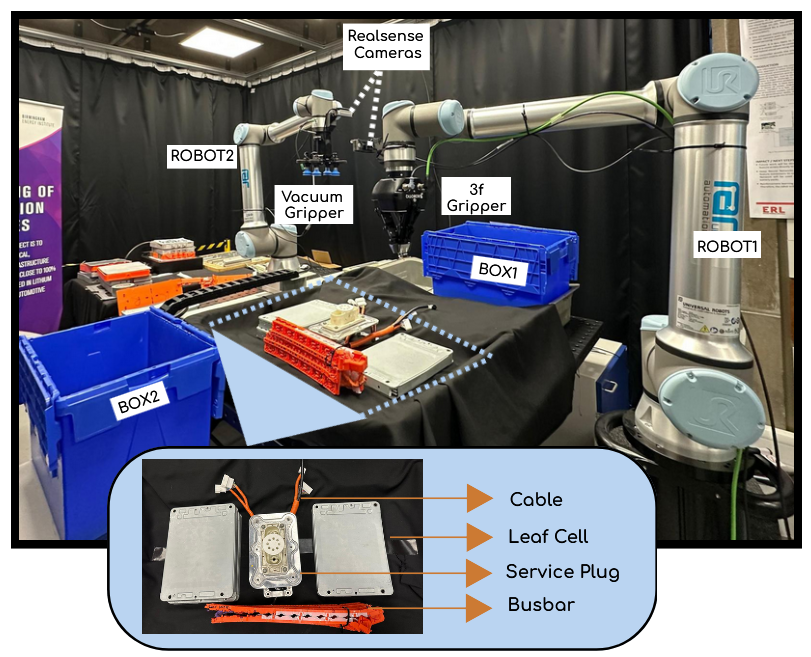}
\caption{The workspace with a battery pack disassembly setup, equipped with two collaborative robot arms with different end-effectors (a vacuum gripper and a 3F gripper) for manipulation tasks and three depth cameras (two cameras for perception and one camera for collision avoidance).} 
\label{case_study_3}
\vspace{-1.5em}
\end{figure}

This case study presents a realistic disassembly scenario in which two UR10e robots collaboratively dismantle components of a NISSAN e-NV200 electric vehicle battery pack. The robots are equipped with distinct end-effectors ($u_1$: vacuum gripper, $u_2$: 3-finger gripper) and are assigned complementary tasks accordingly. The disassembly process involves removing key components such as battery modules, cables, busbars, and a service plug (Fig.~\ref{case_study_3}). Robots operate without prior knowledge of object locations and rely exclusively on vision-based perception, as in our prior work on robotic cutting \cite{rastegarpanah2021vision}. 
\vspace{-1em}

The proposed TAMP framework (Fig.~\ref{fig:TAMP_overview}) is executed here as a closed loop. Perception first localises battery components and confirms their current completion states. Task decomposition and task allocation then generate the current robot-specific action lists according to disassembly order $\pi(\mathcal{O})$, tool requirements, and distance-based cost (Equation~\ref{eq:distance_cost}). Specifically, robot $R_1$, equipped with the 3-finger gripper, manipulates components such as screws, cables, busbars, and service plugs, while robot $R_2$, fitted with the vacuum gripper, handles battery modules and cells. This complementary tool setup enables flexible handling of heterogeneous component geometries while maintaining the expert-defined disassembly order.

The motion layer integrates four TP-GMM models: two for the pick phase (one per robot) and two for the place phase. Each model is trained, as described in Section~\ref{sec:lfd-guided_planner}, using demonstrations tailored to the respective robot and manipulation stage. During task execution, the robots query the TP-GMM server, which returns the appropriate model for the corresponding robot and stage (pick or place). Collision avoidance is embedded within the control strategy. Both robots continuously monitor their shared workspace to detect potential collisions with each other, as well as with static or dynamic objects. In the event of a collision risk, dynamic replanning is triggered (Algorithm~\ref{alg:active_collision}), after which the robots re-scan the workspace and update the remaining task list before committing the next action. The framework executes iteratively until the disassembly sequence is fully completed.
\vspace{-1em}
\subsection{Comparison Metrics}
Each case study was evaluated under both planners, Default and TP-GMM, using \textbf{Path Length} and \textbf{Path Duration}. Study-specific metrics included the \textbf{Manipulability Index} (Case Study II) and \textbf{Swept volume} (Case Study III).
Path length is computed as the cumulative Euclidean distance of end-effector positions along the trajectory between the specified start and end. Reported time metrics denote end-to-end elapsed time from task confirmation in the current scan to successful completion of the corresponding motion. For the proposed planner, this therefore includes TP-GMM reproduction, constrained motion planning, physical execution, and any delays introduced by reactive avoidance/replanning. We report this integrated timing because the contribution of the paper is the coupled real-world system rather than an isolated planner micro-benchmark. Trajectories and cumulative-distance plots are reported for qualitative inspection.

In Case Study II, we additionally evaluate the \textbf{manipulability index}, a configuration-dependent measure of dexterity. It quantifies the arm’s ability to generate end-effector motion across directions, enabling comparison of motion planners and highlighting proximity to singular configurations, particularly informative in cluttered or dynamic scenes. We compute the index following Yoshikawa’s formulation \cite{doi:10.1137/0607034}. In Case Study II, manipulability is evaluated using Yoshikawa’s index, denoted \(w\) \cite{doi:10.1137/0607034}. Joint and end-effector velocities satisfy $\dot{x} = J(q)\dot{q}$ where \(\dot{x}\), \(J(q)\), and \(\dot{q}\) denote the end-effector velocity, the Jacobian at configuration \(q\), and joint velocities, respectively. To characterise directional dexterity, we analyse the eigenvalues of \(JJ^{T}\) and report the condition number

\begin{equation}
    w_{\lambda} = \frac{\lambda_{\max}}{\lambda_{\min}},
\end{equation}

with larger values indicating proximity to singular configurations, thus loss of dexterity.

In Case Study~III, we evaluate the \textbf{swept volume} to quantify the manipulator’s spatial footprint, the 3D region traversed by the links and end-effector over the trajectory. This metric is pertinent in multi-robot or human–robot settings, where lower swept volume implies reduced collision risk and better workspace utilisation. We compute it via a voxel-occupancy model using the same 0.03~m voxel size as the OctoMap collision representation, i.e., the workspace is discretised into cubes of edge \(h=0.03\,\mathrm{m}\), and the total volume is
\begin{equation}
    V = N_{\text{occ}}\, h^{3},
\end{equation}
with \(N_{\text{occ}}\) the number of voxels occupied during motion. This provides a consistent basis for comparing planners in terms of spatial efficiency. Throughout all case studies, the near-collision trigger is fixed at $\varepsilon=0.15$~m, i.e., five occupancy cells, which provides a conservative but practical early-warning margin for reactive execution.
\vspace{-1em}

\section{Results and Discussion}
This section reports the quantitative outcomes of the three case studies introduced in Section~\ref{sec:exp_study}, with emphasis on the contribution-specific metrics of trajectory compactness, reactive safety, and end-to-end coordination quality.
\vspace{-1em}

\subsection{Case Study I: Robot-to-Robot Collision Avoidance}

\begin{figure}[!t]
  \centering
    \subfloat[\centering]{\label{RobotToRobot_path_deviation_a}\includegraphics[width=0.92\columnwidth]{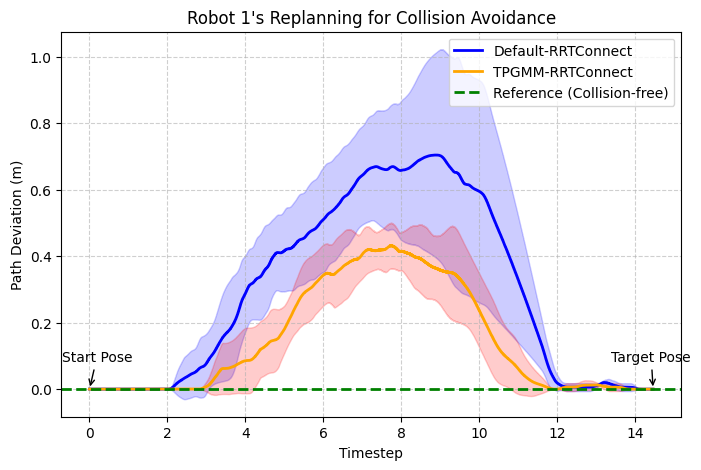}}
    \vspace{-1em}
    \subfloat[\centering]{\label{RobotToRobot_path_deviation_b}\includegraphics[width=0.92\columnwidth]{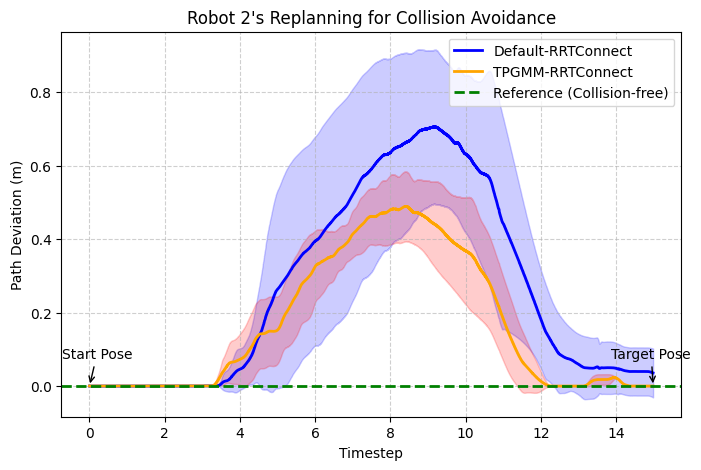}}
  \caption{Comparison of Path Deviations due to collisions detected whilst motion execution of both robots.}
  \label{RobotToRobot_path_deviation}
  \vspace{-1.5em}
\end{figure}

In this case study, we evaluate collision avoidance and re-planning capabilities when two robots operate within a shared workspace. Both robots start from distinct initial poses and move toward predefined targets, creating the potential for inter-robot collisions. The comparison focuses on the deviation from a reference collision-free trajectory, thereby quantifying the quality of the trajectory each planner produces  (Fig.~\ref{RobotToRobot_path_deviation}).  

Fig.~\ref{RobotToRobot_path_deviation_a} shows the deviation profiles of $R_1$. With Default-RRTConnect (blue line), $R_1$ reaches a peak mean deviation of $0.7\,\mathrm{m}$, indicating substantial divergence from the collision-free reference. By contrast, the GMM-Informed RRTConnect planner (orange line) maintains significantly lower deviations, peaking at approximately $0.4\,\mathrm{m}$. This reduction of nearly $45\%$ highlights the stabilizing influence of the learning-informed prior. Similarly, for $R_2$ (Fig.~\ref{RobotToRobot_path_deviation_b}), Default-RRTConnect produces deviations exceeding $0.8\,\mathrm{m}$, while GMM-Informed RRTConnect consistently remains below $0.6\,\mathrm{m}$. Across the trajectories, the shaded regions (variance bands) further highlight the robustness of the GMM-Informed RRTConnect. In particular, Default-RRTConnect exhibits larger variability, with standard deviations of approximately $0.3$ for $R_1$ and $0.25$ for $R_2$, whereas the GMM-Informed RRTConnect maintains a maximum standard deviation of only $0.1$. This reduction in variability indicates more consistent and reliable replanning performance. Both robots converge back to near-zero deviation when reaching their targets, confirming successful task completion. However, the difference lies in the intermediate trajectories: Default-RRTConnect tends to produce abrupt detours with higher variability, whereas GMM-Informed RRTConnect achieves smoother and more predictable paths closer to the reference trajectory (green dashed line).
\vspace{-1em}

\subsection{Case Study II: Dynamic Environment Evaluation}
\label{sec:case_study_II_results}

This case study examines the comparative performance of the Default-RRTConnect and the proposed GMM-Informed RRTConnect planners in a dynamic task execution scenario. The evaluation considers three main aspects: (i) motion planning and execution time, (ii) Cartesian path length, and (iii) manipulability index along the executed trajectories. The goal is to assess the efficiency, consistency, and adaptability of the GMM-Informed planner relative to the baseline.

\begin{figure}[!t]
  \centering
     \subfloat[\centering]{\label{RobotToDynamic1a}\includegraphics[width=0.48\columnwidth]{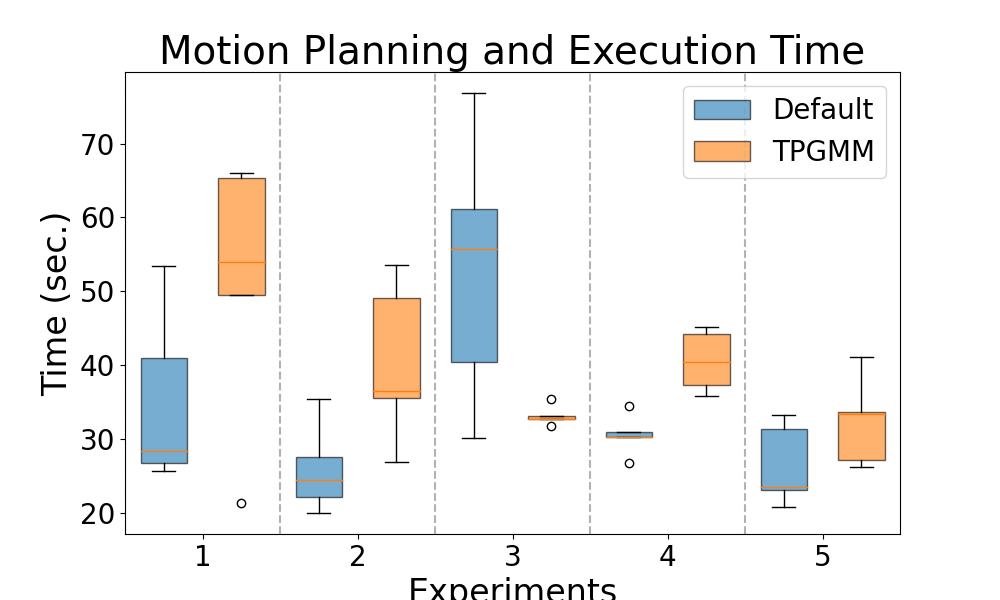}}
    \subfloat[\centering]{\label{RobotToDynamic1b}\includegraphics[width=0.48\columnwidth]{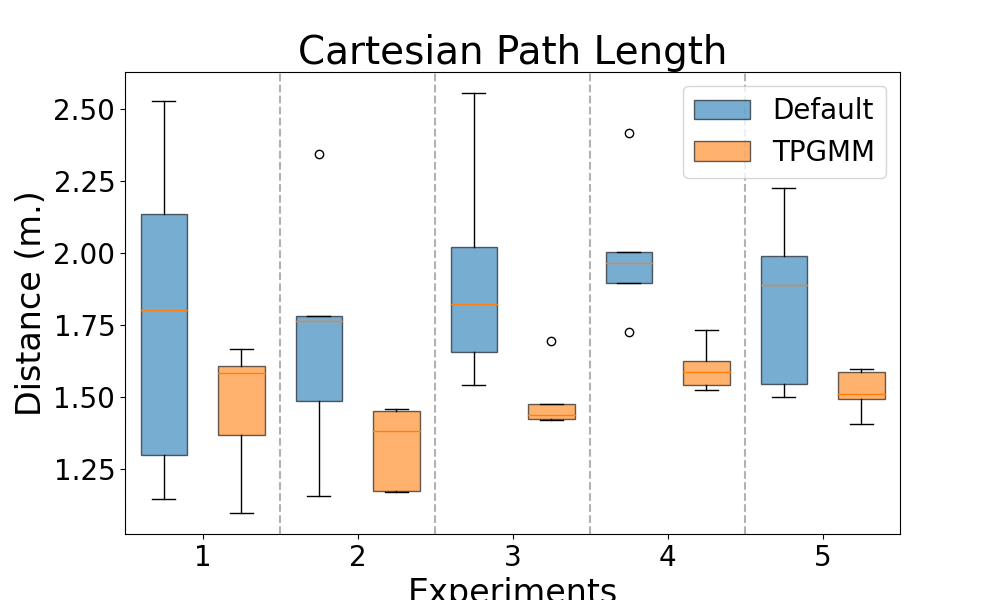}}
  \caption{Comparison of RRTConnect and GMM-Informed planner in terms of (a) end-to-end motion planning and execution time, and (b) Cartesian path length.}
  \label{RobotToDynamic1}
  \vspace{-1.5em}
\end{figure}

Fig. \ref{RobotToDynamic1} presents the experimental outcomes for two of these performance indicators across five independent experiments, each repeated 5 times. Subfigure (a) illustrates motion planning and execution times, while Subfigure (b) depicts the resulting Cartesian path lengths.
In terms of execution time (Fig. \ref{RobotToDynamic1}a), the GMM-Informed planner generally requires more time than the Default-RRTConnect. This additional time is due to the reproduction step of the TP-GMM, which generates an expert-informed trajectory in task space once queried with new task parameters (start, goal, and obstacle poses), as described in Equation \ref{eq:reproduced_gmm}. This reproduced trajectory is then imposed as a constraint in the sampling process, biasing the planner toward expert-like solutions but at the cost of added computation. Despite this increase in planning time, the observed trade-off is justified by the improved quality of the generated trajectories.
When comparing Cartesian path lengths (Fig. \ref{RobotToDynamic1}b), the GMM-Informed planner consistently produces shorter and more efficient paths across all experiments. The Default-RRTConnect, on the other hand, often yields longer and more variable paths, reflecting its tendency to explore larger portions of the configuration space without the benefit of task-informed guidance. The reduced variance in path length observed for the GMM-Informed planner further highlights its robustness and consistency in dynamic settings.
These results indicate that while the GMM-Informed planner introduces a moderate computational overhead, it offers significant improvements in trajectory optimality and reliability; two critical aspects in dynamic, multi-robot disassembly scenarios.

\begin{figure}[!t]
  \centering
    \subfloat[\centering ]{\label{ManipulabilityIndex1a}\includegraphics[width=0.48\columnwidth]{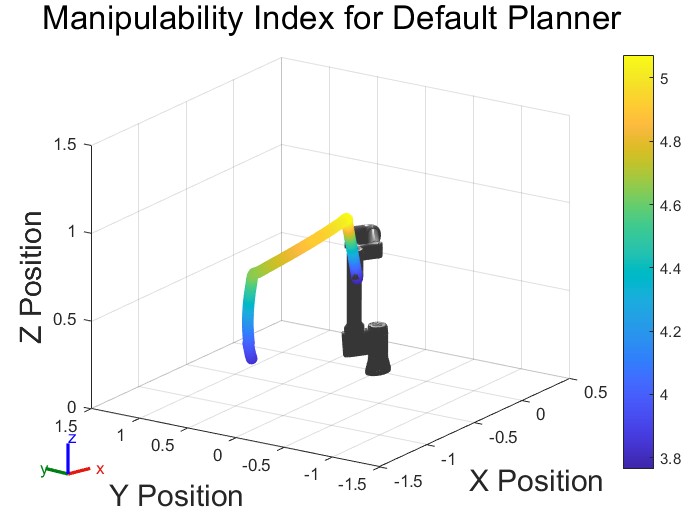}}
    \subfloat[\centering]{\label{ManipulabilityIndex2b}\includegraphics[width=0.48\columnwidth]{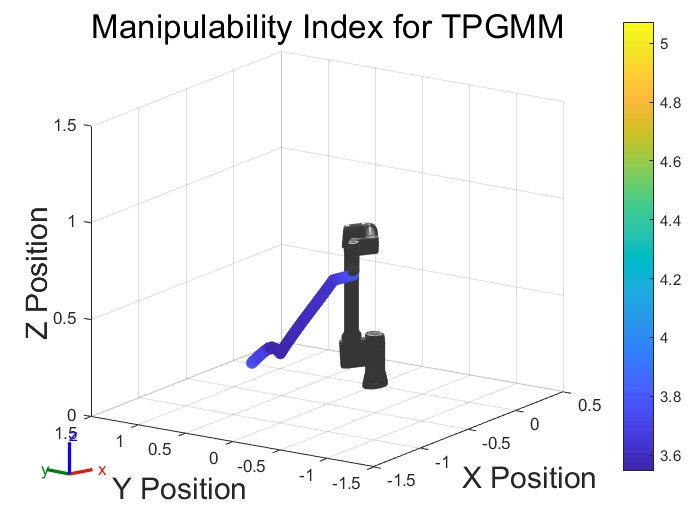}}
  \caption{Comparison of manipulability Index for two trajectory solutions produced with (a) RRTConnect and (b) GMM-informed planner.}
  \label{ManipulabilityIndex}
  \vspace{-1.5em}
\end{figure}

Moreover, as shown in Fig. \ref{ManipulabilityIndex}(a), a Default-RRTConnect trajectory exhibits noticeable spikes in the condition-number-based manipulability metric. These spikes indicate that the robot passes close to singular configurations, where joint mobility becomes restricted and control may be less stable. Conversely, a trajectory generated by the GMM-Informed planner (Fig. \ref{ManipulabilityIndex}(b)) demonstrates a more consistent manipulability profile with lower variance, avoiding abrupt changes that could compromise motion stability.
Overall, while the Default planner occasionally attains locally favourable postures, it does so at the risk of approaching singular configurations. In contrast, the GMM-Informed planner produces smoother, more reliable trajectories that better preserve dexterity, making it a safer choice for dynamic and sensitive multi-robot tasks.
\vspace{-1em}


\subsection{Comparative Evaluation of GMM-Informed and Default Planner in Robotic Disassembly}
\label{subsec:planner_comparison}
Two UR10e robots perform a collaborative EV-battery disassembly sequence in a shared cell (workspace and cell layout in Fig.~\ref{case_study_3}). We compare a learning-based RRTConnect guided by TP-GMM against the Default-RRTConnect planner. Performance is assessed by (i) disassembly time per object and overall makespan (a proxy for coordination efficiency), (ii) end-effector path length (trajectory efficiency), and (iii) swept-volume overlap (spatial footprint in the shared workspace). Both planners execute under the same task allocation and perception pipeline (Section \ref{sec:task_planner}). 
Table~\ref{tab:performance_comparison} summarises the results of 5 trials recorded for each planner in one disassembly environment setup. The number of real-world experiments remains limited in this case study as each dual-arm disassembly run is costly, experimentally complex, and time-intensive to record, monitor, and reset. Moreover, given that the disassembly task sequence is fixed by the look-up table (discussed in section \ref{sec:lookup_table}), five trials were considered sufficient to demonstrate the framework's behaviour. Averaged over all six objects, the GMM-Informed planner reduces cumulative end-effector path length from $48.8 \pm 2.7$\,m to $17.9 \pm 0.9$\,m (\(-63.3\%\)), indicating substantially more compact motions in the shared workspace. It also improves the overall makespan from $467.9 \pm 16.4$\,s to $429.8 \pm 37.1$\,s, despite the added TP-GMM reproduction step. A two-sided Welch t-test on the 5 trial aggregates confirms that the path-length reduction is statistically significant ($t=24.28$, $p<0.001$), whereas the makespan improvement remains a positive trend but is not significant at this sample size ($t=2.10$, $p=0.085$). This result is consistent with the role of the proposed planner: its strongest measurable benefit is producing more compact, lower-interference motions, while end-to-end cycle time additionally depends on task interleaving and reactive pauses.

\begin{table*}[!t]
\centering
\begin{tabular}{|c|c|cc|cc|}
\hline
\multirow{2}{*}{\textbf{Robot}} & \multirow{2}{*}{\textbf{Object Name}} & \multicolumn{2}{c|}{\textbf{RRTConnect}} & \multicolumn{2}{c|}{\textbf{GMM-Informed}} \\
\cline{3-6}
& & \textbf{Time (s)} & \textbf{Path Length (m)} & \textbf{Time (s)} & \textbf{Path Length (m)} \\
\hline
\multirow{3}{*}{Robot 1} 
& Cable         & $79.5 \pm 19.4$ & $7.0 \pm 2.2$ & $86.8 \pm 10.6$ & $3.5 \pm 0.4$ \\
& Busbar        & $80.7 \pm 12.9$ & $8.4 \pm 3.2$ & $85.3 \pm 21.2$ & $3.1 \pm 0.7$ \\
& Service Plug  & $75.9 \pm 13.1$ & $7.9 \pm 0.6$ & $63.3 \pm 17.7$ & $2.3 \pm 0.1$ \\
\hline
\multirow{3}{*}{Robot 2} 
& LeafCell1         & $76.5 \pm 19.9$ & $9.0 \pm 2.4$ & $67.9  \pm 8.5$ & $3.4 \pm 0.8$ \\
& LeafCell2         & $73.1 \pm 16.7$ & $8.4 \pm 2.3$ & $55.8  \pm 5.1$ & $2.4 \pm 0.4$ \\
& LeafCell3         & $82.3 \pm 18.3$ & $7.9 \pm 2.4$ & $70.6 \pm 14.2$ & $3.1 \pm 0.4$ \\
\hline
& \textbf{Disassembly Time:} & $467.9 \pm 16.4$ & & $429.8 \pm 37.1$ & \\
\hline
& \textbf{Path Length:} &  & $48.8 \pm 2.7$ & & $17.9 \pm 0.9$ \\
\hline
\end{tabular}
\caption{Performance comparison of RRTConnect and GMM-informed planner for multi-robot electric-vehicle battery disassembly. Values are reported as mean $\pm$ standard deviation over 5 trials.}
\label{tab:performance_comparison}
\vspace{-1.5em}
\end{table*}

\begin{figure}[ht]
\centering
\subfloat[\centering ]
{\label{fig:swept_volume_compare-a} \includegraphics[width=0.47\columnwidth]{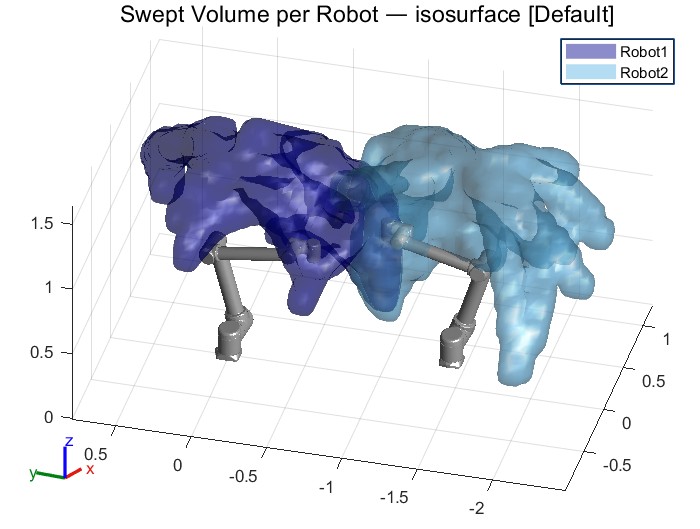}} 
\subfloat[\centering ]
{\label{fig:swept_volume_compare-b} \includegraphics[width=0.47\columnwidth]{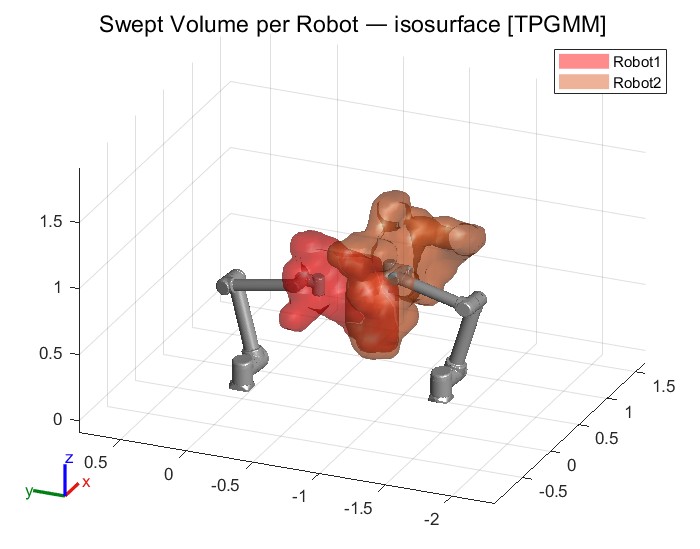}}
\caption{Swept-volume isosurfaces computed on the 0.03~m occupancy grid. (a) RRTConnect and (b) GMM-informed planner.}
\label{fig:swept_volume_compare}
\vspace{-1em}
\end{figure}

In terms of per-object trajectory quality, the GMM-informed planner consistently shortens paths across all objects (approximate reductions: Cable \(\sim\!50\%\), Busbar \(\sim\!63\%\), Service Plug \(\sim\!71\%\), LeafCell1 \(\sim\!62\%\), LeafCell2 \(\sim\!71\%\), LeafCell3 \(\sim\!61\%\)). Timing shows complementary strengths: the learning-based planner is faster for the Service Plug and all LeafCell removals (\(\approx 12\)-\(24\%\) faster), whereas the Default planner is slightly faster on Cable and Busbar (\(\approx 6\)-\(9\%\)), where fewer collision-driven detours occur. This behaviour reinforces the main contribution of the proposed integration: predictive, demonstration-informed planning is most beneficial when dual-arm interaction and dynamic collision handling matter most.
Furthermore, isosurfaces reconstructed from the swept volumes (Fig.~\ref{fig:swept_volume_compare}) reveal that the GMM-Informed planner constrains the arms to a smaller portion of the workspace and handles collision instances more efficiently. Relative to Default, the swept volume drops from \(V_1=0.583~\mathrm{m}^3\) to \(0.139~\mathrm{m}^3\) for Robot~1 (\(-76\%\)) and from \(V_2=0.696~\mathrm{m}^3\) to \(0.252~\mathrm{m}^3\) for Robot~2 (\(-64\%\)). The mutual overlap volume decreases from \(0.064~\mathrm{m}^3\) to \(0.034~\mathrm{m}^3\) (\(-47\%\)), showing that the combined predictive/reactive strategy not only shortens paths but also reduces workspace sharing conflicts during concurrent manipulation.
\vspace{-1.5em}

Although the overall results are favourable, several typical failure cases were observed during experimentation. First, the framework remains sensitive to perception outages and occlusions. If a target component becomes temporarily occluded from one or both cameras, or if a camera stream fails during execution, the system may continue searching for the object that is next in the disassembly sequence. In the current implementation, such failures are not always recoverable online and may require restarting the full task. Second, deadlock-like behaviour can arise when one robot executes an aggressive collision-avoidance manoeuvre and approaches a near-singular configuration. In this case, the corresponding local planner may become effectively stuck, preventing progress while the other robot waits for the shared workspace to clear. Third, the learned TP-GMM guidance is less reliable when the robot starts from a configuration already close to singularity, because the demonstrated motion priors are not sufficiently representative of those low-dexterity regions. In practice, however, this issue occurs only rarely in our experiments. These observations highlight the need for stronger perception-failure recovery, explicit deadlock-resolution policies, and singularity-aware learning or initialisation strategies in future versions of the system.

\vspace{-1em}

\section{Conclusions}
This paper presented a perception-driven Task-and-Motion Planning (TAMP) framework for multi-robot EV battery disassembly. The key contribution is the coordination between three essential elements on a real dual-arm platform: (i) task decomposition and task allocation driven by repeated scene perception, (ii) the recently introduced GMM-informed RRT motion planner \cite{SHAARAWY2026103095}, and (iii) a hybrid predictive/reactive collision-handling strategy that combines digital-twin validation with online avoidance and replanning. The challenge is not merely assembling these modules, but making them operate jointly under real-world uncertainty, dual-arm coordination complexity, dynamic obstacles, and simultaneous task-level and geometric constraints.

At the task level, stereo vision with YOLOv8 and depth sensing detects, segments, and localises components so that logical disassembly constraints can be enforced. An expert-encoded lookup table then drives task decomposition and task allocation while remaining extensible to new components and related sequential tasks. Importantly, the system does not treat task planning and motion planning as fully decoupled: after each action, the robots re-scan the workspace, update completion status, and recompute the remaining task sequence when necessary.

For motion planning, demonstrations are processed into Task-Parametrised Gaussian Mixture Models (TP-GMMs) that guide feasible, expert-like trajectories. Predictive collision checking through OctoMap, MoveIt, and FCL is combined with reactive steering and replanning during execution. Across the three case studies, this integrated formulation yields more compact and safer dual-arm behaviour. In the end-to-end disassembly scenario, average path length decreases from \(48.8\,\mathrm{m}\) to \(17.9\,\mathrm{m}\) (\(-63.3\%\)), with a statistically significant improvement over Default-RRTConnect ($p<0.001$), while makespan improves from \(467.9\,\mathrm{s}\) to \(429.8\,\mathrm{s}\) (\(-8.1\%\)) with higher variability across the 5 trials. Swept-volume analysis further shows substantially smaller per-arm volumes ($R1$: \(0.583 \rightarrow 0.139\,\mathrm{m}^3\), $R2$: \(0.696 \rightarrow 0.252\,\mathrm{m}^3\)) and reduced overlap (\(0.064 \rightarrow 0.034\,\mathrm{m}^3\)), supporting the claim that predictive planning plus reactive collision avoidance improves workspace sharing in real dual-arm disassembly.

Beyond the specific EV-disassembly use case, the modular design of this framework makes it adaptable to larger robot teams and other sequential manipulation tasks. A current limitation is that dual-arm coupling is learned only indirectly through shared scene monitoring and replanning; jointly learned dual-arm motion priors remain an important direction for future work. Overall, the results show that the proposed integration is both practically useful and technically meaningful for coordinated multi-robot manipulation in unstructured and dynamically changing environments.

\vspace{-1em}
\section*{Funding}
This work was funded by the project called “Research and Development of a Highly Automated and Safe Streamlined Process for Increasing Lithium-ion Battery Repurposing and Recycling” (REBELION) under Grant 10079049, and partially supported by the Ministry of National Education, Republic of Turkey.
\vspace{-1.5em}



 
%

\bibliography{refs}

@article{562456321,
author={Xidias,E. and Moulianitis,V. and Azariadis,P.},
year={2021},
month={07},
title={Optimal robot task scheduling based on adaptive neuro-fuzzy system and genetic algorithms},
journal={The International Journal of Advanced Manufacturing Technology},
volume={115},
number={3},
pages={927-939},
note={Copyright - © Springer-Verlag London Ltd., part of Springer Nature 2020; Last updated - 2023-11-25},
isbn={02683768},
language={English},
}

@article{FATEMIANARAKI2023102770,
title = {Scheduling of Multi-Robot Job Shop Systems in Dynamic Environments: Mixed-Integer Linear Programming and Constraint Programming Approaches},
journal = {Omega},
volume = {115},
pages = {102770},
year = {2023},
issn = {0305-0483},
doi = {https://doi.org/10.1016/j.omega.2022.102770},
author = {Soroush Fatemi-Anaraki and Reza Tavakkoli-Moghaddam and Mehdi Foumani and Behdin Vahedi-Nouri}
}

@ARTICLE{9681247,
  author={Touzani, Hicham and Séguy, Nicolas and Hadj-Abdelkader, Hicham and Suárez, Raúl and Rosell, Jan and Palomo-Avellaneda, Leopold and Bouchafa, Samia},
  journal={IEEE Robotics and Automation Letters}, 
  title={Efficient Industrial Solution for Robotic Task Sequencing Problem With Mutual Collision Avoidance \& Cycle Time Optimization}, 
  year={2022},
  volume={7},
  number={2},
  pages={2597-2604},
  doi={10.1109/LRA.2022.3142919}}

@article{BERNARDO2023109345,
title = {A novel framework to improve motion planning of robotic systems through semantic knowledge-based reasoning},
journal = {Computers \& Industrial Engineering},
volume = {182},
pages = {109345},
year = {2023},
issn = {0360-8352},
doi = {https://doi.org/10.1016/j.cie.2023.109345},
author = {Rodrigo Bernardo and João M.C. Sousa and Paulo J.S. Gonçalves}
}

@article{rastegarpanah2021vision,
  title={Vision-guided mpc for robotic path following using learned memory-augmented model},
  author={Rastegarpanah, Alireza and Hathaway, Jamie and Stolkin, Rustam},
  journal={Frontiers in Robotics and AI},
  volume={8},
  pages={688275},
  year={2021},
  publisher={Frontiers Media SA}
}

@article{WEGENER2014155,
title = {Disassembly of Electric Vehicle Batteries Using the Example of the Audi Q5 Hybrid System},
journal = {Procedia CIRP},
volume = {23},
pages = {155-160},
year = {2014},
note = {5th CATS 2014 - CIRP Conference on Assembly Technologies and Systems},
issn = {2212-8271},
doi = {https://doi.org/10.1016/j.procir.2014.10.098},
author = {Kathrin Wegener and Stefan Andrew and Annika Raatz and Klaus Dröder and Christoph Herrmann}
}

@article{Wang2023ABatReSimAC,
  title={ABatRe-Sim: A Comprehensive Framework for Automated Battery Recycling Simulation},
  author={Huanqing Wang and Kaixiang Zhang and Keyi Zhu and Ziyou Song and Zhaojian Li},
  journal={2023 IEEE International Conference on Robotics and Biomimetics (ROBIO)},
  year={2023},
  pages={1-8},
}

@INPROCEEDINGS{9636119,
  author={Pan, Tianyang and Wells, Andrew M. and Shome, Rahul and Kavraki, Lydia E.},
  booktitle={2021 IEEE/RSJ International Conference on Intelligent Robots and Systems (IROS)}, 
  title={A General Task and Motion Planning Framework For Multiple Manipulators}, 
  year={2021},
  volume={},
  number={},
  pages={3168-3174},
  doi={10.1109/IROS51168.2021.9636119}}

@article{TEJER2024107300,
title = {Robust and efficient task scheduling for robotics applications with reinforcement learning},
journal = {Engineering Applications of Artificial Intelligence},
volume = {127},
pages = {107300},
year = {2024},
issn = {0952-1976},
doi = {https://doi.org/10.1016/j.engappai.2023.107300},
author = {Mateusz Tejer and Rafal Szczepanski and Tomasz Tarczewski}
}

@inbook{inbook,
author = {Wang, Hanfu and Chen, Weidong},
year = {2022},
month = {01},
pages = {276-287},
title = {Simulated Annealing Algorithms for the Heterogeneous Robots Task Scheduling Problem in Heterogeneous Robotic Order Fulfillment Systems},
isbn = {978-3-030-95891-6},
doi = {10.1007/978-3-030-95892-3_21},
publisher = {Springer}
}

@article{WANG2024104604,
author = {Wang, Hanfu and Chen, Weidong},
title = {Task scheduling for heterogeneous agents pickup and delivery using recurrent open shop scheduling models},
journal = {Robotics and Autonomous Systems},
volume = {172},
pages = {104604},
year = {2024},
issn = {0921-8890},
doi = {https://doi.org/10.1016/j.robot.2023.104604},
url = {https://www.sciencedirect.com/science/article/pii/S0921889023002439}
}

@Article{electronics12092131,
AUTHOR = {Bae, Sanghyeon and Joo, Sunghyeon and Choi, Junhyeon and Pyo, Jungwon and Park, Hyunjin and Kuc, Taeyong},
TITLE = {Semantic Knowledge-Based Hierarchical Planning Approach for Multi-Robot Systems},
JOURNAL = {Electronics},
VOLUME = {12},
YEAR = {2023},
NUMBER = {9},
ARTICLE-NUMBER = {2131},
ISSN = {2079-9292},
DOI = {10.3390/electronics12092131}
}

@misc{YOLOV8,
  author    = {Jacob Solawetz, Francesco},
  title     = {What is YOLOv8? The Ultimate Guide},
  year      = {2024},
  note      = {Accessed:5 September 2024},
  howpublished = {\url{https://blog.roboflow.com/whats-new-in-yolov8/}},
}

@article{DBLP:journals/corr/ColemanSCC14,
  author       = {David Coleman and
                  Ioan Alexandru Sucan and
                  Sachin Chitta and
                  Nikolaus Correll},
  title        = {Reducing the Barrier to Entry of Complex Robotic Software: a MoveIt!
                  Case Study},
  journal      = {CoRR},
  volume       = {abs/1404.3785},
  year         = {2014},
  eprinttype    = {arXiv},
  eprint       = {1404.3785},
  bibsource    = {dblp computer science bibliography, https://dblp.org}
}

@INPROCEEDINGS{6225337,
  author={Pan, Jia and Chitta, Sachin and Manocha, Dinesh},
  booktitle={2012 IEEE International Conference on Robotics and Automation}, 
  title={FCL: A general purpose library for collision and proximity queries}, 
  year={2012},
  volume={},
  number={},
  pages={3859-3866},
  doi={10.1109/ICRA.2012.6225337}}

@Article{robotics13050075,
AUTHOR = {Erdogan, Cansu and Contreras, Cesar Alan and Stolkin, Rustam and Rastegarpanah, Alireza},
TITLE = {Multi-Robot Task Planning for Efficient Battery Disassembly in Electric Vehicles},
JOURNAL = {Robotics},
VOLUME = {13},
YEAR = {2024},
NUMBER = {5},
ARTICLE-NUMBER = {75},
ISSN = {2218-6581},
DOI = {10.3390/robotics13050075}
}

@article{doi:10.1137/0607034,
author = {Togai, Masaki},
title = {An Application of the Singular Value Decomposition to Manipulability and Sensitivity of Industrial Robots},
journal = {SIAM Journal on Algebraic Discrete Methods},
volume = {7},
number = {2},
pages = {315-320},
year = {1986},
doi = {10.1137/0607034},
eprint = {https://doi.org/10.1137/0607034}
}

@INPROCEEDINGS{8594492,
  author={McMahon, Troy and Jenkins, Odest Chadwicke and Amato, Nancy},
  booktitle={2018 IEEE/RSJ International Conference on Intelligent Robots and Systems (IROS)}, 
  title={Affordance Wayfields for Task and Motion Planning}, 
  year={2018},
  volume={},
  number={},
  pages={2955-2962},
  doi={10.1109/IROS.2018.8594492}}

@INPROCEEDINGS{8612805,
  author={Zhao, Yingshen and Fillatreau, Philippe and Karray, Mohamed Hedi and Archimede, Bernard},
  booktitle={2018 IEEE/ACS 15th International Conference on Computer Systems and Applications (AICCSA)}, 
  title={An Ontology-Based Approach Towards Coupling Task and Path Planning for the Simulation of Manipulation Tasks}, 
  year={2018},
  volume={},
  number={},
  pages={1-8},
  doi={10.1109/AICCSA.2018.8612805}}

@INPROCEEDINGS{10092974,
  author={Antão, Liliana and Costa, Nuno and Gonçalves, Gil},
  booktitle={2022 2nd International Conference on Robotics, Automation and Artificial Intelligence (RAAI)}, 
  title={General Purpose Task and Motion Planning for Human-Robot Teams}, 
  year={2022},
  volume={},
  number={},
  pages={8-14},
  doi={10.1109/RAAI56146.2022.10092974}}

@ARTICLE{9013090,
  author={Motes, James and Sandström, Read and Lee, Hannah and Thomas, Shawna and Amato, Nancy M.},
  journal={IEEE Robotics and Automation Letters}, 
  title={Multi-Robot Task and Motion Planning With Subtask Dependencies}, 
  year={2020},
  volume={5},
  number={2},
  pages={3338-3345},
  doi={10.1109/LRA.2020.2976329}}

@ARTICLE{508439,
  author={Kavraki, L.E. and Svestka, P. and Latombe, J.-C. and Overmars, M.H.},
  journal={IEEE Transactions on Robotics and Automation}, 
  title={Probabilistic roadmaps for path planning in high-dimensional configuration spaces}, 
  year={1996},
  volume={12},
  number={4},
  pages={566-580},
  doi={10.1109/70.508439}}

@article{LaValle1998RapidlyexploringRT,
  title={Rapidly-exploring random trees : a new tool for path planning},
  author={Steven M. LaValle},
  journal={The annual research report},
  year={1998},
}

@inproceedings{inproceedings,
author = {Thomason, Wil and Knepper, Ross},
year = {2019},
month = {06},
pages = {},
title = {A Unified Sampling-Based Approach to Integrated Task and Motion Planning}
}

@INPROCEEDINGS{9911132,
  author={Realpe, Sebastian and Roldan, Felipe Gonzalez and Fajardo, Jose Manuel and Hernández, Juan D. and Cardenas, Pedro-F.},
  booktitle={2022 27th International Conference on Automation and Computing (ICAC)}, 
  title={Benchmark of Sampling Based Motion Planners in Bin Picking Manipulation Task}, 
  year={2022},
  volume={},
  number={},
  pages={1-6},
  doi={10.1109/ICAC55051.2022.9911132}}

@misc{ortizharo2024idbrrtsamplingbasedkinodynamicmotion,
      title={iDb-RRT: Sampling-based Kinodynamic Motion Planning with Motion Primitives and Trajectory Optimization}, 
      author={Joaquim Ortiz-Haro and Wolfgang Hönig and Valentin N. Hartmann and Marc Toussaint and Ludovic Righetti},
      year={2024},
      eprint={2403.10745},
      archivePrefix={arXiv},
      primaryClass={cs.RO}, 
}

@misc{khanal2023guidedsamplingbasedmotionplanning,
      title={Guided Sampling-Based Motion Planning with Dynamics in Unknown Environments}, 
      author={Abhish Khanal and Hoang-Dung Bui and Gregory J. Stein and Erion Plaku},
      year={2023},
      eprint={2306.09229},
      archivePrefix={arXiv},
      primaryClass={cs.RO},
}

@article{SHAARAWY2026103095,
title = {Task-aware motion planning in constrained environments using GMM-informed RRT planners},
journal = {Robotics and Computer-Integrated Manufacturing},
volume = {97},
pages = {103095},
year = {2026},
issn = {0736-5845},
doi = {https://doi.org/10.1016/j.rcim.2025.103095},
author = {Abdelaziz Shaarawy and Alireza Rastegarpanah and Rustam Stolkin}
}

@article{TeleOp_paper,
  title={Towards reuse and recycling of lithium-ion batteries: Tele-robotics for disassembly of electric vehicle batteries},
  author={Hathaway, Jamie and Shaarawy, Abdelaziz and Akdeniz, Cansu and Aflakian, Ali and Stolkin, Rustam and Rastegarpanah, Alireza},
  journal={Frontiers in Robotics and AI},
  volume={10},
  pages={1179296},
  year={2023},
  publisher={Frontiers Media SA}
}

@INPROCEEDINGS{9613554,
  author={Sarwar, Muhammad Usman and Sohail, Moman and Din, Muhayy Ud and Rosell, Jan and Qazi, Wajahat M},
  booktitle={2021 26th IEEE International Conference on Emerging Technologies and Factory Automation (ETFA )}, 
  title={A Dataset Generation Tool for Deep learning-based Motion Planning in Complex Environments}, 
  year={2021},
  volume={},
  number={},
  pages={1-4},
  doi={10.1109/ETFA45728.2021.9613554}}

@Article{machines11070722,
AUTHOR = {Noroozi, Fatemeh and Daneshmand, Morteza and Fiorini, Paolo},
TITLE = {Conventional, Heuristic and Learning-Based Robot Motion Planning: Reviewing Frameworks of Current Practical Significance},
JOURNAL = {Machines},
VOLUME = {11},
YEAR = {2023},
NUMBER = {7},
ARTICLE-NUMBER = {722},
ISSN = {2075-1702},
DOI = {10.3390/machines11070722}
}

@article{Fan_2023,
doi = {10.1088/1742-6596/2580/1/012048},
year = {2023},
month = {sep},
publisher = {IOP Publishing},
volume = {2580},
number = {1},
pages = {012048},
author = {Fan, Zeyu},
title = {Multi-Point path planning for robots based on deep reinforcement learning},
journal = {Journal of Physics: Conference Series}
}

@ARTICLE{hornung13auro,
  author = {Armin Hornung and Kai M. Wurm and Maren Bennewitz and Cyrill
  Stachniss and Wolfram Burgard},
  title = {{OctoMap}: An Efficient Probabilistic {3D} Mapping Framework Based
  on Octrees},
  journal = {Autonomous Robots},
  year = 2013,
  doi = {10.1007/s10514-012-9321-0},
}

@INPROCEEDINGS{10802695,
  author={Pan, Tianyang and Verginis, Christos K. and Kavraki, Lydia E.},
  booktitle={2024 IEEE/RSJ International Conference on Intelligent Robots and Systems (IROS)}, 
  title={Robust and Safe Task-Driven Planning and Navigation for Heterogeneous Multi-Robot Teams with Uncertain Dynamics}, 
  year={2024},
  volume={},
  number={},
  pages={3482-3489},
  doi={10.1109/IROS58592.2024.10802695}}

@INPROCEEDINGS{10802307,
  author={Bouhsain, Smail Ait and Alami, Rachid and Siméon, Thierry},
  booktitle={2024 IEEE/RSJ International Conference on Intelligent Robots and Systems (IROS)}, 
  title={Extending Task and Motion Planning with Feasibility Prediction: Towards Multi-Robot Manipulation Planning of Realistic Objects}, 
  year={2024},
  volume={},
  number={},
  pages={10318-10325},
  doi={10.1109/IROS58592.2024.10802307}}

@ARTICLE{10807203,
  author={Hathaway, Jamie and Contreras, Cesar Alan and Asif, Mohammed Eesa and Stolkin, Rustam and Rastegarpanah, Alireza},
  journal={IEEE Access}, 
  title={Technoeconomic Assessment of Electric Vehicle Battery Disassembly-Challenges and Opportunities From a Robotics Perspective}, 
  year={2025},
  volume={13},
  number={},
  pages={716-733},
  doi={10.1109/ACCESS.2024.3520414}}

@article{sucan2012the-open-motion-planning-library,
    Author = {Ioan A. {\c{S}}ucan and Mark Moll and Lydia E. Kavraki},
    Doi = {10.1109/MRA.2012.2205651},
    Journal = {{IEEE} Robotics \& Automation Magazine},
    Month = {December},
    Number = {4},
    Pages = {72--82},
    Title = {The {O}pen {M}otion {P}lanning {L}ibrary},
    Volume = {19},
    Year = {2012}
}
\bibliographystyle{IEEEtran}

\end{document}